\documentclass[11pt]{article}

\usepackage[utf8]{inputenc}
\usepackage[T1]{fontenc}
\usepackage{lmodern}
\usepackage[margin=1in]{geometry}
\usepackage[authoryear,round]{natbib}
\usepackage{amsmath,amssymb,amsfonts,amsthm,mathtools}
\usepackage{booktabs}
\usepackage{graphicx}
\usepackage{flafter}
\usepackage[section]{placeins}
\usepackage{tikz}
\usetikzlibrary{arrows.meta,calc,fit,positioning,shapes.geometric}
\usepackage{microtype}
\usepackage{xcolor}
\usepackage{hyperref}
\usepackage{url}

\definecolor{darkblue}{rgb}{0.0,0.0,0.65}
\definecolor{darkred}{rgb}{0.68,0.05,0.0}
\hypersetup{
  colorlinks=true,
  citecolor=darkblue,
  linkcolor=darkred,
  filecolor=darkblue,
  urlcolor=darkblue,
  pdftitle={On Sparse Graph Limits for Neural Combinatorial Optimization},
  pdfauthor={Thien Le and Melanie Weber}
}

\newtheorem{theorem}{Theorem}
\newtheorem{proposition}[theorem]{Proposition}
\newtheorem{lemma}[theorem]{Lemma}
\newtheorem{corollary}[theorem]{Corollary}
\newtheorem{definition}[theorem]{Definition}
\newtheorem{remark}[theorem]{Remark}

\newcommand{\cC}{\mathcal{C}}
\newcommand{\cK}{\mathcal{K}}
\newcommand{\cQ}{\mathcal{Q}}
\newcommand{\cS}{\mathcal{S}}

\title{On Sparse Graph Limits for Neural Combinatorial Optimization}
\author{
Thien Le\\
SEAS, Harvard University\\
Cambridge, MA\\
\texttt{thien\_le@seas.harvard.edu}
\and
Melanie Weber\\
SEAS, Harvard University\\
Cambridge, MA\\
\texttt{mweber@seas.harvard.edu}
}
\date{}

\begin{document}
\maketitle

\begin{abstract}
Graph neural networks are often trained on small graphs and deployed on larger
ones, raising the question of which graph information must be preserved for
combinatorial optimization values to transfer across sizes. We establish
size-transfer guarantees for normalized graph optimization values using finite
collections of energy features. For affine objectives, including MAX-CUT,
coloring defect, and independent-set density, differences in task value are
controlled by discrepancies in normalized scalar energy probes on all
bounded-degree graphs. For general Lipschitz objectives, such as those with
balance penalties, an additional representation error appears; this error
vanishes on hyperfinite graph families but can remain bounded away from zero
on expanders. Thus, at fixed degree bound, state space, Lipschitz constant, and
target accuracy, a finite energy representation independent of graph size
suffices universally for affine tasks and asymptotically for general
Lipschitz tasks on hyperfinite families. These results yield prediction
guarantees under explicit conditions on encoder accuracy, readout stability,
and representation coverage, together with statistical generalization bounds.
Theoretically, we prove Shapley--Folkman bounds with sharp parameter dependence,
including a dimension-free $\ell_2$ regime, and realize the resulting energy
features with a fixed-weight message-passing network using random node
initialization on bounded-treewidth graphs. Experiments with exact solvers and
trained models test the resulting transfer guarantees.

\end{abstract}

\section{Introduction}
\label{sec:introduction}

Graph neural networks are routinely trained on small graphs and deployed on much larger ones, raising the question of when predictions can reliably transfer across sizes. Four lines of work address different aspects of this question.  Work on \emph{size
transferability} bounds how much the output of
a fixed model changes as the input graph grows, for spectral filters
and dense and sparse graph limits
\citep{levie2019transferability,levie2021transferability,ruiz2020graphon,
keriven2020convergence,maskey2023transferability,ruiz2023transferability,
cervino2023transference,le2023graphoptransfer}, with a parallel empirical line on size
extrapolation
\citep{yehudai2021size,bevilacqua2021size}.  {Work on
\emph{statistical generalization} bounds the gap between training and test error, whether
through classical capacity measures
\citep{garg2020generalization,liao2021pacbayes,morris2023wlvc,
vasileiou2025covered}, random-graph models
\citep{maskey2022generalization}, or graphon, attributed-graph, graphop, and
manifold limit spaces
\citep{levie2023graphon,rauchwerger2025attributed,amran2026graphop,
wang2025manifold}.  Related work studies universality and expressivity on
large random graphs \citep{keriven2021universality,keriven2023functions}.  A
\emph{signal-processing} line studies sampling and filtering on finite graphs,
graphons, and manifolds
\citep{chamon2018greedy,ruiz2021graphonsp,ruiz2021architectures,
wang2022manifolds,wang2024stability,wang2024geometric,le2024poincare,
paradamayorga2024sampling}.}  Finally, \emph{neural combinatorial
optimization} trains networks to output relaxed solutions to hard graph
problems
\citep{bengio2021machine,cappart2023combinatorial,karalias2020erdos,
toenshoff2021maxcsp,schuetz2022combinatorial,wang2022unsupervised,
karalias2022extensions,wang2023meta,sanokowski2024diffusion,vanderbush2026neural}, for which
small-to-large deployment is standard but solver reliability remains debated
\citep{angelini2023modern}.

This paper connects these questions for combinatorial optimization on arbitrarily large graphs. In problems such as coloring, MAX-CUT, and maximum independent set, the objective is local but the optimum depends on a global assignment and cannot be recovered from bounded-radius neighborhoods. We ask which graph information must be retained for task values to transfer across sizes. Too coarse a representation precludes architecture-independent guarantees; too fine a representation retains information the task never uses.

Bounded-degree graph limit theory supplies a candidate representation.
For fixed $q$ (e.g. number of colors in coloring), one $q$-state assignment produces a quotient point: it
records the fraction of vertices in each state and the normalized edge
frequency between every pair of states.  The \emph{$q$-quotient set}
$\cS_q(G)$ collects these aggregate profiles over all assignments (see
Figure~\ref{fig:quotient-construction}).  Hausdorff closeness of two
quotient sets means that every profile attainable on either graph has a
nearby counterpart on the other.  Requiring this for every fixed $q$ is
\emph{partition convergence} \citep{bollobas2011sparse,borgs2017ld}, so
continuous objectives of quotient statistics have stable optima across
graph sizes.

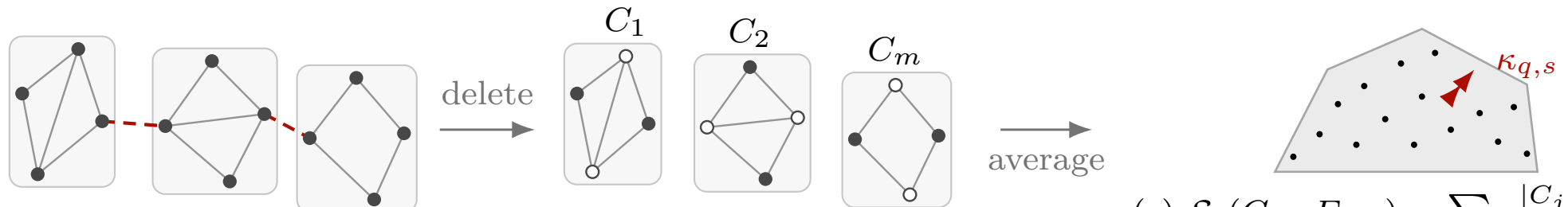
\begin{figure}[ht]
\centering
\begin{tikzpicture}[xscale=.99,
  >=Latex,
  every node/.style={font=\scriptsize},
  vertex/.style={circle,fill=black!72,inner sep=1.15pt},
  openvertex/.style={circle,draw=black!72,fill=white,
    line width=.45pt,inner sep=1.05pt},
  edge/.style={draw=black!42,line width=.45pt},
  cut/.style={draw=darkred,densely dashed,line width=.8pt},
  region/.style={draw=black!22,fill=black!3,
    rounded corners=3pt,inner sep=3pt}]

\begin{scope}[xshift=0cm]
  \coordinate (a1) at (-1.25,.45); \coordinate (a2) at (-.78,.82);
  \coordinate (a3) at (-.58,.22);  \coordinate (a4) at (-1.12,-.22);
  \coordinate (b1) at (-.05,.18);  \coordinate (b2) at (.34,.72);
  \coordinate (b3) at (.78,.28);   \coordinate (b4) at (.49,-.28);
  \coordinate (c1) at (1.16,.08);  \coordinate (c2) at (1.55,.58);
  \coordinate (c3) at (1.95,.12);  \coordinate (c4) at (1.70,-.43);
  \node[region,fit=(a1)(a2)(a3)(a4)] {};
  \node[region,fit=(b1)(b2)(b3)(b4)] {};
  \node[region,fit=(c1)(c2)(c3)(c4)] {};
  \draw[edge] (a1)--(a2)--(a3)--(a4)--(a1) (a2)--(a4);
  \draw[edge] (b1)--(b2)--(b3)--(b4)--(b1) (b1)--(b3);
  \draw[edge] (c1)--(c2)--(c3)--(c4)--(c1);
  \draw[cut] (a3)--(b1) (b3)--(c1);
  \foreach \v in {a1,a2,a3,a4,b1,b2,b3,b4,c1,c2,c3,c4}
    \node[vertex] at (\v) {};
  \node at (.35,-.88)
    {(a) $G$, with $|E_{\rm cut}|\leq\varepsilon n$};
\end{scope}

\draw[->,black!55,line width=.55pt] (2.25,.15)--(3.05,.15)
  node[midway,above=2pt] {delete};

\begin{scope}[xshift=4.55cm]
  \coordinate (d1) at (-1.15,.45); \coordinate (d2) at (-.74,.76);
  \coordinate (d3) at (-.55,.20);  \coordinate (d4) at (-1.02,-.20);
  \coordinate (e1) at (-.06,.17);  \coordinate (e2) at (.30,.67);
  \coordinate (e3) at (.70,.25);   \coordinate (e4) at (.43,-.26);
  \coordinate (f1) at (1.18,.07);  \coordinate (f2) at (1.52,.52);
  \coordinate (f3) at (1.88,.10);  \coordinate (f4) at (1.64,-.39);
  \node[region,fit=(d1)(d2)(d3)(d4)] {};
  \node[region,fit=(e1)(e2)(e3)(e4)] {};
  \node[region,fit=(f1)(f2)(f3)(f4)] {};
  \draw[edge] (d1)--(d2)--(d3)--(d4)--(d1) (d2)--(d4);
  \draw[edge] (e1)--(e2)--(e3)--(e4)--(e1) (e1)--(e3);
  \draw[edge] (f1)--(f2)--(f3)--(f4)--(f1);
  \foreach \v in {d1,d3,e2,e4,f1,f3}
    \node[vertex] at (\v) {};
  \foreach \v in {d2,d4,e1,e3,f2,f4}
    \node[openvertex] at (\v) {};
  \node[above=1pt] at (d2) {$C_1$};
  \node[above=1pt] at (e2) {$C_2$};
  \node[above=1pt] at (f2) {$C_m$};
  \node at (.35,-.88)
    {(b) $G-E_{\rm cut}=\bigsqcup_j C_j$, $|C_j|\leq K$};
\end{scope}

\draw[->,black!55,line width=.55pt] (6.95,.15)--(7.72,.15)
  node[midway,below=2pt] {average};

\begin{scope}[xshift=9.25cm,yshift=-.20cm,x=2.2cm,y=1.25cm]
  \fill[black!8]
    (0,0)--(.20,.68)--(.56,.95)--(.96,.58)--(1,0)--cycle;
  \draw[black!35,line width=.5pt]
    (0,0)--(.20,.68)--(.56,.95)--(.96,.58)--(1,0)--cycle;
  \foreach \p in {
    (.07,.10),(.17,.25),(.24,.45),(.31,.18),
    (.34,.57),(.42,.35),(.48,.72),(.53,.18),
    (.56,.50),(.61,.79),(.67,.28),(.71,.58),
    (.78,.39),(.85,.20),(.91,.43),(.96,.12)}
    \fill \p circle (.75pt);
  \draw[<->,darkred,line width=.55pt] (.71,.58)--(.76,.68);
  \node[darkred,anchor=west] at (.78,.70) {$\kappa_{q,s}$};
  \node[align=center] at (.52,-.28)
    {(c) $\cS_q(G-E_{\rm cut})
      =\sum_j\frac{|C_j|}{n}\cS_q(C_j)$};
  \node[align=center] at (.52,-.53)
    {dots $\longrightarrow$ shaded convex hull};
\end{scope}
\end{tikzpicture}
\caption{Hyperfinite decomposition and convexification.  In a hyperfinite
family, with $n:=|V(G)|$, deleting $E_{\rm cut}$,
$|E_{\rm cut}|\leq\varepsilon n$, leaves components
$C_1,\ldots,C_m$ with $|C_j|\leq K(\varepsilon)$ and $K$ independent of $n$.
Independent component assignments make
$\cS_q(G-E_{\rm cut})$ a vertex-weighted Minkowski sum (dots);
Shapley--Folkman bounds its gap $\kappa_{q,s}$ to the shaded convex hull.}
\label{fig:decomposition-mechanism}
\end{figure}

Scalar energy features offer a simpler representation by optimizing
affine scores over $\cS_q(G)$. But since these quotient sets are nonconvex in general, we can optimize instead over $\operatorname{conv}\cS_q(G)$ and employ tools from convex optimization.
However, convexification introduces mixtures
of profiles that need not come from any single assignment.  How close
 $\cS_q(G)$ is to $\operatorname{conv}\cS_q(G)$ is therefore a worst-case rounding, or
representation, error: convexification leaves affine optima unchanged,
but contributes an additional transfer error for nonlinear Lipschitz\footnote{Lipschitzness is defined by equipping the partition convergence topology with an $\ell^s$ metric, for $s \in \{1,2\}$}
tasks.  Figure~\ref{fig:decomposition-mechanism} shows the mechanism that
makes this error small on hyperfinite graphs while Table~\ref{tab:transfer-regimes} states the four
resulting transfer regimes.

\begin{table}[ht]
\caption{Task-value transfer by regime.  Here
$D$ is the maximum-degree bound; $T_F(X):=\operatorname{opt}_{z\in\cS_q(X)}F(z)$,
where $\operatorname{opt}$ is the task's minimum or maximum; $L_s$ is the
Lipschitz constant under the $\ell^s$ quotient metric; $\eta_{q,s}$ is the
$\ell^s$-Hausdorff distance between the convexified sets;
$\kappa_{q,s}(X)$ is the set-to-hull distance; and $n_X:=|V(X)|$.
Under an $(\varepsilon_X,K_X)$ decomposition,
$\rho_1=O_D(\varepsilon_X+q^2K_X/n_X)$ and
$\rho_2=O_D(\varepsilon_X+\min\{qK_X/n_X,\sqrt{K_X/n_X}\})$. Note that the regime $\sqrt{K_X/n_X}$ is independent of $q$}
\label{tab:transfer-regimes}
\centering
\footnotesize
\setlength{\tabcolsep}{3pt}
\renewcommand{\arraystretch}{1.06}
\begin{tabular}{@{}p{.29\linewidth}p{.25\linewidth}p{.40\linewidth}@{}}
\toprule
Task class & Graph assumption & Bound on $|T_F(G)-T_F(H)|$ \\
\midrule
Affine: MAX-CUT, coloring defect, independent-set density
 & Any $D$-bounded $G,H$ & $L_s\eta_{q,s}(G,H)$ \\
General $L_s$-Lipschitz, including finite balance penalties
 & Any $D$-bounded $G,H$
 & $L_s[\eta_{q,s}+\kappa_{q,s}(G)+\kappa_{q,s}(H)]$ \\
Same general class, $s=1$
 & Both graphs admit such decompositions
 & $L_1[\eta_{q,1}+\rho_1(G)+\rho_1(H)]$ \\
Same general class, $s=2$
 & Same
 & $L_2[\eta_{q,2}+\rho_2(G)+\rho_2(H)]$ \\
\bottomrule
\end{tabular}
\end{table}


The $q$- and $K/n$-dependences of the $\ell^1$ rate, and both branches of
the $\ell^2$ rate, are separately sharp; the optimal $\varepsilon$
coefficient remains open (Appendix~\ref{sec:sharpness}).
For a concrete example, appendix~\ref{app:example_intro} works through the affine/nonlinear
distinction and these bounds for coloring.

\paragraph{Contributions.}
For finite simple graphs of maximum degree at most a fixed $D$, we obtain:

\begin{enumerate}
    \item \textbf{Transfer of task values.}
    Differences in a finite set of normalized energy features control differences in affine task values on all graphs. For general Lipschitz objectives, an additional representation error appears, quantified in Table~\ref{tab:transfer-regimes}
    (Theorem~\ref{thm:task-value-size-transfer}).

    \item \textbf{When energy features are enough.}
    We show using Shapley--Folkman arguments that this additional error vanishes on hyperfinite graph sequences
    (Theorem~\ref{thm:hyperfinite-convexification} and
    Corollary~\ref{cor:hyperfinite-scalar-completeness}), at essentially sharp rates
    (Remark~\ref{thm:sharp-convexification}). Outside this regime, a construction based on expanders gives an $\Omega(1)$ separation
    (Proposition~\ref{prop:expander-separation}).

    \item \textbf{Consequences for learning.}
    These results yield prediction and generalization guarantees that do not depend on graph size
    (Theorem~\ref{thm:task-value-size-transfer},
    Proposition~\ref{prop:probe-generalization}). On bounded-treewidth graphs, the required energy features can be computed by a fixed-weight message-passing network with random node initialization, while high-girth constructions show that logarithmic depth is necessary in general
    (Theorems~\ref{thm:raw-rni-encoder}--\ref{thm:rni-depth-lower}).
\end{enumerate}

\subsection{Related work}
\label{sec:related-work}

We cover only the closest related works here; Appendix~\ref{app:extended-related}
gives the full graph-limit genealogy and theorem-level distinctions.

\textbf{Quotient convergence.}
Our $(p,W)$ coordinates and Hausdorff convergence of all $\cS_q(G_n)$ are
exactly the partition convergence of \citet{borgs2017ld}.  Dense and
uniformly upper-regular theories connect quotients to constrained energies
\citep{bclsv2012,borgs2018lp2}, but their normalization or regularity
excludes bounded degree.  At the vertex scale, unconstrained energies fail
in general; Corollary~\ref{cor:hyperfinite-scalar-completeness} identifies
the hyperfinite regime in which they suffice.

\textbf{Convexification.}
Shapley--Folkman bounds for separable nonconvex programs are classical
\citep{starr1969quasi,aubin1976estimates,udell2016bounding}.  We specialize
them to quotient sets, prove matching lower bounds, and identify the
associated Lagrangian gap (Remark~\ref{thm:sharp-convexification} and
Proposition~\ref{prop:lagrangian-bridge}).

\textbf{Graph learning and neural optimization.}
Size generalization asks when models trained on small graphs predict on
larger ones \citep{yehudai2021size}; we ask which task information must be
retained.  Neural relaxations produce fractional assignments, whereas
overlap-gap barriers concern their optimization
\citep{karalias2020erdos,schuetz2022combinatorial,gamarnik2017limits}.
Our transfer and rounding bounds isolate the preceding representation
question from that optimization question.

\section{Quotient sets and partition convergence}
\label{sec:attainable-regions}

To study transfer across graph sizes, we first need a representation that characterizes exactly the information relevant to the task. We summarize each graph by the statistics produced by all possible assignments, collected in a \emph{quotient set}, and compare these summaries across graph sizes using \emph{partition convergence}.

Fix a nonnegative integer $D$. Throughout, $G$ is a finite, simple, nonempty
graph with vertex set $V:=V(G)$, edge set $E:=E(G)$, maximum degree
$\Delta(G)\leq D$, and $n:=|V(G)|$ vertices; $[m]:=\{1,\ldots,m\}$ for every
positive integer $m$, $q$ is the state-space size, and
$\vec E(G):=\{(u,v):uv\in E(G)\}$ is the oriented edge set.

\begin{figure}[!htb]
\centering
\begin{tikzpicture}[>=Latex,every node/.style={font=\scriptsize}]
\begin{scope}[xshift=0cm]
 \foreach \i/\a/\s in {1/90/1,2/162/1,3/234/2,4/306/1,5/18/2}{
   \coordinate (v\i) at (\a:0.75);
 }
 \draw[black!60] (v1)--(v2)--(v3)--(v4)--(v5)--(v1);
 \foreach \i/\s in {1/1,2/1,4/1}{\fill (v\i) circle (2.8pt);}
 \foreach \i/\s in {3/2,5/2}{\draw[fill=white] (v\i) circle (2.8pt);}
 \node at (0,-1.45) {(a) $c$ on $C_5$};
 \node at (0,-1.85) {\tiny$\bullet=$ state $1$, $\circ=$ state $2$};
\end{scope}
\node at (1.75,0) {$\Rightarrow$};
\begin{scope}[xshift=3.65cm]
 \draw[black!60] (-0.7,0)--(0.7,0);
 \draw[black!60] (-0.7,0.1) .. controls (-1.3,0.8) and (-0.1,0.8) .. (-0.7,0.1);
 \fill (-0.7,0) circle (3.6pt);
 \draw[fill=white] (0.7,0) circle (3.6pt);
 \node at (-1.0,-0.42) {$p_1=\frac35$};
 \node at (1.0,-0.42) {$p_2=\frac25$};
 \node at (0,-0.95) {$W_{12}=W_{21}=\frac45$};
 \node at (-0.7,1.05) {$W_{11}=\frac25$};
 \node at (0,-1.5) {(b) $G/c=(p,W)$};
 \node at (0,-1.9) {\tiny$W_{22}=0$};
\end{scope}
\node at (5.55,0) {$\Rightarrow$};
\begin{scope}[xshift=6.5cm,yshift=-1.0cm,x=1.85cm,y=1.28cm]
 \draw[->,black!60,thin] (0,0)--(1.1,0) node[right] {$p_1$};
 \draw[->,black!60,thin] (0,0)--(0,1.12) node[above] {$W_{12}$};
 \foreach \p in {(0,0),(0.2,0.4),(0.4,0.4),(0.4,0.8),(0.6,0.4),(0.8,0.4),(1,0)}
   {\fill \p circle (1.5pt);}
 \fill (0.6,0.8) circle (1.5pt);
 \draw (0.6,0.8) circle (3.4pt);
 \node at (0.55,-0.28) {(c) $\cS_2(C_5)$, $G/c$ circled};
\end{scope}
\begin{scope}[xshift=9.75cm,yshift=-1.0cm,x=1.85cm,y=1.28cm]
 \fill[black!8] (0,0)--(0.5,1)--(1,0)--cycle;
 \draw[black!30,thin] (0,0)--(0.5,1)--(1,0)--cycle;
 \draw[->,black!60,thin] (0,0)--(1.1,0) node[right] {$p_1$};
 \draw[->,black!60,thin] (0,0)--(0,1.12) node[above] {$W_{12}$};
 \foreach \p in {(0,0),(0.1,0.2),(0.2,0.2),(0.2,0.4),(0.3,0.2),(0.3,0.4),(0.3,0.6),(0.4,0.2),(0.4,0.4),(0.4,0.6),(0.4,0.8),(0.5,0.2),(0.5,0.4),(0.5,0.6),(0.5,0.8),(0.5,1),(0.6,0.2),(0.6,0.4),(0.6,0.6),(0.6,0.8),(0.7,0.2),(0.7,0.4),(0.7,0.6),(0.8,0.2),(0.8,0.4),(0.9,0.2),(1,0)}
   {\draw \p circle (1.05pt);}
 \foreach \x in {1,...,29}{
   \pgfmathsetmacro\xm{\x/30}
   \pgfmathtruncatemacro\rmax{min(\x,30-\x)}
   \foreach \r in {1,...,\rmax}{
     \pgfmathsetmacro\ym{2*\r/30}
     \fill (\xm,\ym) circle (0.5pt);
   }
 }
 \fill (0,0) circle (0.5pt); \fill (1,0) circle (0.5pt);
 \node at (0.55,-0.28) {(d) $\cS_2(C_{10})\ \circ$, $\cS_2(C_{30})\ \cdot$};
\end{scope}
\end{tikzpicture}
\caption{The fundamental objects on a worked example (exact rational
coordinates).  (a)~An assignment $c=(1,1,2,1,2)$ on $C_5$.  (b)~Its
$q$-quotient $G/c=(p,W)$ (Definition~\ref{def:q-quotient}).  (c)~The
quotient set $\cS_2(C_5)$ in the $(p_1,W_{12})$-projection, $G/c$ circled.
(d)~Partition convergence: $\circ$ marks $\cS_2(C_{10})$, $\cdot$ marks
$\cS_2(C_{30})$; the denser grid approaches the shaded limiting region
(exact formula: Appendix~\ref{app:example_quotient_set}).}
\label{fig:quotient-construction}
\end{figure}
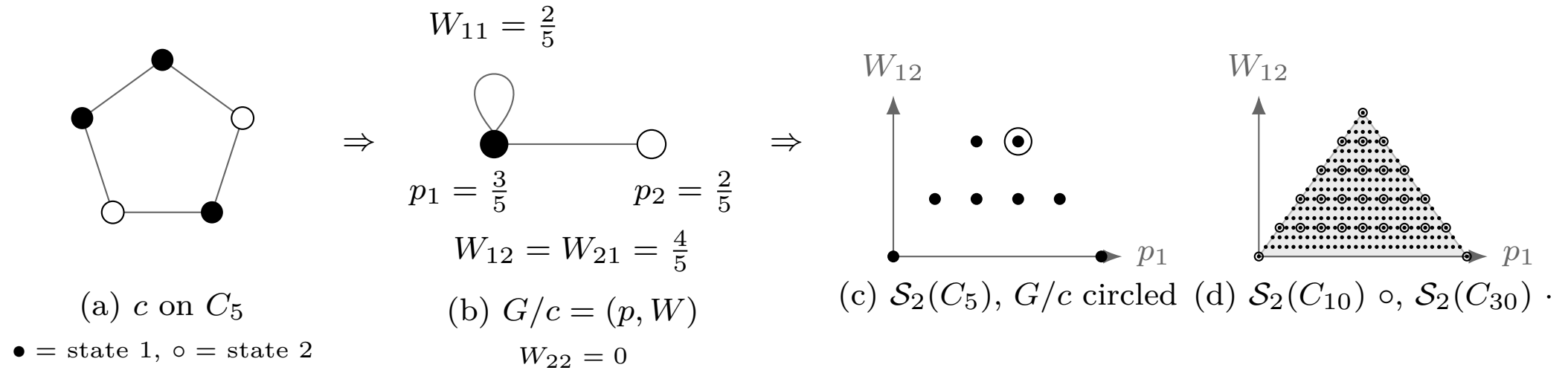

\begin{definition}[$q$-quotient and quotient set]
\label{def:q-quotient}
For an arbitrary assignment $c:V(G)\to[q]$, define its node weights $p=(p_i)_{i=1}^q$ and oriented edge weights
$W=(W_{ij})_{i,j=1}^q$ by
\begin{equation}
 p_i(G,c)
 :=\frac1n\#\{v:c(v)=i\},
 \qquad
 W_{ij}(G,c)
 :=\frac1n\#\{(u,v)\in\vec E(G):c(u)=i,\ c(v)=j\}.
 \label{eq:vertex-proportion}
\end{equation}
 The weighted $q$-vertex graph $G/c\equiv (p(G,c), W(G,c))$ is the
\emph{$q$-quotient} induced by $c$.  The \emph{quotient set} of $G$ is the collection of all quotients induced by some assignment
\begin{equation}
 \cS_q(G):=\{G/c \mid c:V(G)\to[q]\}.
 \label{eq:quotient-set}
\end{equation}
\end{definition}
Figure~\ref{fig:quotient-construction}(a)--(c) traces one assignment into its
quotient point and then into the full quotient set.  Vertex
normalization makes the quotient well defined for edgeless graphs and
linear under vertex-weighted disjoint unions. Moreover, each quotient satisfies
$W=W^\top$, $\sum_{i,j}W_{ij}=2|E(G)|/n\leq D$, and
$\sum_jW_{ij}\leq Dp_i$ for all $i$.  With
$\Delta_q:=\{p\in[0,1]^q:\sum_ip_i=1\}$ the simplex, quotients live in
the metric space
\begin{equation}
 \begin{aligned}
 \mathsf Y_{D,q}
 &:=\bigl\{(p,W):p\in\Delta_q,\ W\in[0,D]^{q\times q},\
 W=W^\top,\ \textstyle\sum_jW_{ij}\leq Dp_i\bigr\},\\
 d_q\bigl((p,W),(p',W')\bigr)
 &:=\|p-p'\|_1+\|W-W'\|_1.
 \end{aligned}
 \label{eq:quotient-host}
\end{equation}
Here and below, the $\ell^1$ norm of a matrix is the entrywise norm.

Thus each quotient set $\cS_q(G)$ is a nonempty compact subset of
$\mathsf Y_{D,q}$, and quotient sets are compared in the Hausdorff metric
$d_{\rm H}(A,B):=\max\{\sup_{a\in A}\inf_{b\in B}d_q(a,b),\,
\sup_{b\in B}\inf_{a\in A}d_q(a,b)\}$
(permutation invariance, worked examples: Appendices~\ref{app:example_quotient_set}
and~\ref{app:deferred-attainable}).

\begin{definition}[Partition convergence]
\label{def:partition-convergence}
A sequence $(G_n)$ of $D$-bounded graphs is \emph{partition convergent} if
$\cS_q(G_n)$ converges in Hausdorff distance for every positive integer $q$.
\end{definition}

To package all state-space sizes, let $\cK(Y)$ denote the hyperspace of
nonempty compact subsets of a compact metric space $Y$, and let
$\mathfrak X_D^{\mathrm{part}}$ be the closure of
$\{(\cS_q(G))_{q\geq1}:\Delta(G)\leq D\}$ in the product
$\prod_{q\geq1}\cK(\mathsf Y_{D,q})$ of Hausdorff hyperspaces.
This \emph{finite-realizable limit space} is compact and metrizable
(Theorem~\ref{thm:compactness}, Appendix~\ref{app:deferred-attainable}), so
every sequence of finite $D$-bounded graphs has a partition-convergent
subsequence (Figure~\ref{fig:quotient-construction}(d)). Thus the limit
state is available without a structural assumption on the graph sequence.
In Appendix~\ref{app:quotient_in_local_global_action}
we give comparisons between quotients and the richer local--global and action topologies.

Intuitively speaking, we can afford to only retain information on aggregate node and edge weights because the tasks that we are interested in are \emph{locally-scored}. The next section identifies which tasks need the full
quotient set and which can be recovered from scalar energies alone.
\FloatBarrier

\section{Continuous quotient problems and ground-state energies}
\label{sec:task-calculus}

We first express
finite-state pairwise energies in the quotient coordinates, then identify exactly
what scalar optimization (e.g. min coloring defect) preserves.

A homogeneous $q$-state pairwise spin model is specified by a unary cost
$a\in\mathbb R^q$ and a symmetric interaction matrix
$B\in\mathbb R^{q\times q}$.  Angle brackets denote Euclidean pairings; in
particular, we write $\langle a,p\rangle:=\sum_i a_ip_i$ and
$\langle B,W\rangle:=\sum_{i,j}B_{ij}W_{ij}$.  For every assignment $c$, its
energy per vertex factors as
\begin{equation}
 \frac1n\left(
   \sum_{v\in V(G)}a_{c(v)}
   +\sum_{uv\in E(G)}B_{c(u)c(v)}
 \right)
 =\langle a,p(G,c)\rangle
  +\frac12\langle B,W(G,c)\rangle.
 \label{eq:pairwise-factorization}
\end{equation}
Minimizing this
quantity is the model's ground-state energy.

\begin{samepage}
\paragraph{Example: Potts antiferromagnet and proper $q$-colorings}\label{ex:potts}
For $a=0$ and the $q\times q$ identity matrix $B=I_q$,
\eqref{eq:pairwise-factorization} is the density
of monochromatic edges---the Hamiltonian of the $q$-state Potts
antiferromagnet---whose ground-state energy is $0$ exactly when
$\chi(G)\le q$, where $\chi(G)$ is the chromatic number
\citep{fortuin1972,sokal2005}. Appendix~\ref{app:deferred-task}
expands the example.  In general, the
next theorem says the quotient supports every continuous post-processing,
and that its topology is the coarsest one that does so. A proof can be found in Appendix~\ref{app:proof-task-completeness}.
\end{samepage}

\begin{theorem}[Completeness of continuous functions over quotient weights]
\label{thm:task-completeness}
Fix $q$, and let $K_n$ and $K$ be nonempty compact subsets of the host
$\mathsf Y_{D,q}$ with $d_{\rm H}(K_n,K)\to0$.  Then for every continuous map
$F:\mathsf Y_{D,q}\to\mathbb R^k$,
\begin{equation}
 d_{\rm H}\bigl(F(K_n),F(K)\bigr)\longrightarrow0.
 \label{eq:continuous-range-transfer}
\end{equation}
In particular, for real-valued $F$,
$\min_{z\in K_n}F(z)\to\min_{z\in K}F(z)$ and
$\max_{z\in K_n}F(z)\to\max_{z\in K}F(z)$.
Conversely, $K_n\to K$ in Hausdorff distance if and only if
$\min_{K_n}F\to\min_KF$ for every real-valued continuous $F$ on
$\mathsf Y_{D,q}$: the Hausdorff topology is the coarsest one on quotient
sets that preserves every continuous query of quotient statistics.
\end{theorem}
In plainer terms, convergence of quotient sets transfers every task value
expressible as a continuous optimum, which is the property a
generalization theory for learned graph invariants must certify.

\subsection{The nonconvex geometry of quotient sets}

When the output is a single scalar affine optimum, the nonconvex geometry
is redundant: the \emph{convexified quotient set}
$\cC_q(G):=\operatorname{conv}\cS_q(G)$ leaves every support function
$h_{\cS_q(G)}(\theta):=\max_{z\in\cS_q(G)}\langle\theta,z\rangle$
unchanged, and the next proposition shows this loss is exact.
We write $\kappa_q(G):=d_{\rm H}(\cS_q(G),\cC_q(G))$ for the resulting
convexification gap.

\begin{proposition}[Scalar ground-state information]
\label{prop:convex-sufficiency}
The optimum values of all homogeneous additive $q$-state unary--pairwise
energies determine exactly $\cC_q(G)$.  For a sequence, convergence of all
support functions is equivalent to Hausdorff convergence of these convexified
quotient sets.
\end{proposition}

The full family of unconstrained scalar energies thus determines the hull
$\cC_q(G)$ and no finer information about $\cS_q(G)$. Consequently every
\emph{affine} task transfers on every graph at a rate controlled by scalar
energy discrepancy (e.g. Corollary~\ref{cor:opening-task-transfer}). The rest of
this section concerns tasks that read the full geometry. One instance when this distinction vanishes asymptotically is for hyperfinite graphs.

\textbf{Hyperfinite graphs.}
The hull nevertheless becomes exact asymptotically when a graph decomposes
into many small pieces.  A sequence is \emph{growing} if
$|V(G_n)|\to\infty$.  A growing
bounded-degree sequence is \emph{hyperfinite}
\citep{schramm2008hyperfinite,newman2013hyperfinite,hatami2014localglobal}
if for every $\varepsilon>0$, deleting at most $\varepsilon|V(G_n)|$
edges leaves components of order at most some $K(\varepsilon)$ independent
of $n$. Paths, cycles, bounded-degree planar, and fixed-minor-free families
are standard examples \citep{newman2013hyperfinite}, and the next theorem
quantifies the cost of convexification under such a decomposition.
\begin{theorem}[Hyperfinite convexification in $\ell^1$]
\label{thm:hyperfinite-convexification}
Suppose that deleting at most $\varepsilon n$ edges from a $D$-bounded graph
$G$ leaves components of order at most $K$.  For every fixed $q$,
\begin{equation}
 d_{\rm H}\!\left(\cS_q(G),\operatorname{conv}\cS_q(G)\right)
 \leq 4\varepsilon
 +\frac{2(1+D)r_qK}{n}, \qquad r_q:=(q-1)+\frac{q(q+1)}2-1.
 \label{eq:hyperfinite-convexification}
\end{equation}
Consequently, if $(G_n)$ is a growing hyperfinite sequence of
$D$-bounded graphs, then the left-hand side tends to zero for every fixed
$q$.
\end{theorem}
Thus the hull is an asymptotically lossless summary.  Whereas component
histograms already determine hyperfinite graphs up to $\varepsilon$ edits
\citep{hassidim2009local,newman2013hyperfinite}, the theorem gives an explicit
$K/n$ rate from unconstrained energies alone.  Its proof combines the
component quotient sets by a weighted Minkowski average and applies
Shapley--Folkman (see Appendix~\ref{app:proof-hyperfinite-convexification}). A concrete
$O_q(n^{-1/2})$ rate for paths and cycles and the theorem's qualitative
relatives are in Appendix~\ref{app:deferred-task}.

\begin{corollary}[Scalar energies suffice on hyperfinite sequences]
\label{cor:hyperfinite-scalar-completeness}
For a growing hyperfinite $D$-bounded sequence $(G_n)$, the
following are equivalent: (1) $(G_n)$ is partition convergent; (2)
 for every finite $q$, every homogeneous $q$-state
 ground-state energy converges.
\end{corollary}

Thus scalar energies suffice on hyperfinite families.  The sharpness remark
and expander counterexample below delineate this conclusion.

\begin{remark}[Sharpness and norm dependence]
\label{thm:sharp-convexification}
The two finite-size terms in Theorem~\ref{thm:hyperfinite-convexification}
are separately sharp: matchings force $\Theta(\min\{1,q^2/n\})$, and unions
of $K$-vertex expanders force $\Omega(K/n)$ already for $q=2$.  Under the
Euclidean quotient metric used in Section~\ref{sec:size-transfer}, the
convexification-gap bound improves to
$O_D(\varepsilon+\min\{qK/n,\sqrt{K/n}\})$, with both branches optimal up to
constants. Interestingly, the second branch operates independently of $q$. Thus $\ell^1$ and $\ell^2$ have the same fixed-$q$ topology but
different uniform finite-$q$ rates. Exact statements and proofs are in
Appendices~\ref{app:deferred-task} and~\ref{app:proof-sharp-convexification}.
\end{remark}

\begin{proposition}[Balanced-cut separation; informal]
\label{prop:expander-separation}
Outside hyperfinite families, a natural constraint can survive in the full
quotient set while disappearing from every scalar energy
\citep[cf.][Example~4]{borgs2017ld}.  Write
$h(G):=\min_{0<|S|\leq|V(G)|/2}|E(S,V(G)\setminus S)|/|S|$ for edge
expansion.  If $h(G_n)\geq h>0$ and $H_n:=G_n\sqcup G_n$, their convexified $q$-quotient sets are
identical for every $q$, so all scalar energies agree.  Yet the balanced-cut
penalty
$\Phi(G):=\min_{(p,W)\in\cS_2(G)}(|p_1-\tfrac12|+W_{12})$ satisfies
$\Phi(H_n)=0$ and $\Phi(G_n)\geq\min\{1/2,h/2\}$.  Hence energy features can
lose a constrained task such as balanced partitioning
\citep{nazi2019gap,gatti2022partitioning}, even though affine tasks still
transfer (an explicit pair is in Appendix~\ref{app:why-convexity}).
\end{proposition}

Proposition~\ref{prop:expander-separation} identifies exactly what scalar energies
lose: the scalar-versus-constrained gap is a Lagrangian duality gap, bounded
by Theorem~\ref{thm:hyperfinite-convexification}.
\begin{proposition}[Lagrangian bridge; informal, with formal statement in Appendix~\ref{app:deferred-task}]
\label{prop:lagrangian-bridge}
On every finite $D$-bounded graph, the gap between a Lipschitz
balance-penalized optimum and its best unconstrained Lagrangian approximation
lies between zero and a Lipschitz multiple of $\kappa_q(G)$.
\end{proposition}

Each Lagrangian approximation is an unconstrained energy shifted by a
constant, so constrained optima are recoverable from implementable
measurements up to the convexification error and no further
(Appendices~\ref{app:deferred-task} and~\ref{app:why-convexity}; proof in
Appendix~\ref{app:proof-lagrangian-bridge}).
Scalar energies are thus complete on hyperfinite sequences, with matching
rates in $K/n$. Outside that regime the failure is an explicit, certified
duality gap.  The next section instantiates the calculus for representative repulsive tasks.

\section{Coloring and compact spin spaces}
\label{sec:examples}

We now express representative repulsive objectives in quotient weights.
Coloring and cuts penalize similar adjacent states, independent set combines
an edge penalty with a vertex reward, and compact spins provide a continuous
relaxation.

\textbf{Coloring.} Put
$u_q(p,W):=\frac12\sum_{i=1}^qW_{ii}$.  For a graph $G$ define its minimum
normalized $q$-coloring defect by
\begin{align}
 e_q(G) :=\frac{1}{n}\min_{c:V(G)\to[q]}
   \#\{uv\in E(G):c(u)=c(v)\}
 =\min_{(p,W)\in\cS_q(G)}u_q(p,W).
 \label{eq:coloring-defect}
\end{align}
Equivalently, $e_q(G)$ is the minimum number of edge deletions per vertex
making $G$ $q$-colorable and Theorem~\ref{thm:task-completeness} extends it
continuously to $\mathfrak X_D^{\mathrm{part}}$.

\textbf{Target-graph colorings and cuts.} Let $H$ be a fixed nonempty loopless graph with vertex set $[h]$.  The
normalized number of violated edges in an $H$-coloring is
\begin{equation}
 \operatorname{defect}_H(G)
 :=\frac1n\min_{\phi:V(G)\to[h]}
 \#\{uv\in E(G):\phi(u)\phi(v)\notin E(H)\}
 =\min_{(p,W)\in\cS_h(G)}
 \frac12\sum_{ij\notin E(H)}W_{ij}.
 \label{eq:target-defect}
\end{equation}
Because this objective is affine in $W$, it is continuous under partition
convergence.  The complementary maximization is MAX $H$-COL
\citep{farnqvist2008hcol}, and taking $H=K_q$, the complete graph on $q$
vertices, gives MAX-$q$-CUT, so that
$\operatorname{MaxCut}(G)/n=\max_{(p,W)\in\cS_2(G)}W_{12}$.

\textbf{Independent sets.}
If state $1$ means ``selected,'' rewarding state-$1$ mass and penalizing
state-$1$ edges yields the exact independence ratio, where $\alpha(G)$
denotes the maximum independent-set size,
$\alpha(G)/n=\max_{(p,W)\in\cS_2(G)}(p_1-\tfrac12W_{11})$ since removing one
endpoint per selected internal edge proves equality.  Hence both the
independence ratio and normalized minimum vertex-cover size are continuous.

\textbf{Compact-spin ground-state energies.}
Finite state spaces also approximate compact spin spaces, uniformly over
graph size thanks to the degree bound.

\begin{proposition}[Compact-spin ground-state transfer]
\label{prop:compact-spin}
Let $K$ be a nonempty compact metric spin space and let
$h:K^2\to\mathbb R$ be a continuous symmetric interaction.  Then
$
 \mathsf E^-_{K,h}(G)
 :=\frac1n\min_{z\in K^{V(G)}}\sum_{uv\in E(G)}h(z_u,z_v)
$
extends continuously to $\mathfrak X_D^{\mathrm{part}}$.  The analogous
maximum extends continuously as well.
\end{proposition}

The finite-state calculus therefore also covers continuous compact-spin
energies (proof in
Appendix~\ref{app:proof-compact-spin}; the unit-sphere example behind the
Lov\'asz $\vartheta$ function is in Appendix~\ref{app:deferred-examples}), and
each example above is a normalized optimum of a Lipschitz function of the
quotient---the shape of task the next section transfers across graph orders.

\section{Task-value transfer across graph sizes}
\label{sec:size-transfer}

The preceding examples are vertex-normalized, so their values compare
across graph orders; we now make the comparison quantitative.
Write $d_{q,1}:=d_q$ and
$d_{q,2}((p,W),(p',W'))
:=(\|p-p'\|_2^2+\|W-W'\|_{\rm F}^2)^{1/2}$, where $\|\cdot\|_{\rm F}$ is
the Frobenius norm, and let $d_{{\rm H},s}$ be the
Hausdorff metric induced by $d_{q,s}$.  For two $D$-bounded graphs $G,G'$,
put, for $s\in\{1,2\}$,
\begin{equation}
 \begin{aligned}
 \delta_{q,s}(G,G')&:=d_{{\rm H},s}(\cS_q(G),\cS_q(G')),\\
 \eta_{q,s}(G,G')&:=d_{{\rm H},s}
   (\operatorname{conv}\cS_q(G),\operatorname{conv}\cS_q(G')),\\
 \kappa_{q,s}(G)&:=d_{{\rm H},s}
   (\cS_q(G),\operatorname{conv}\cS_q(G)).
 \end{aligned}
 \label{eq:normwise-discrepancies}
\end{equation}
If
$\theta=(a,B)$, where $a\in\mathbb R^q$ and $B$ is symmetric, write
$\langle B,W\rangle:=\sum_{i,j}B_{ij}W_{ij}$ and
$\|B\|_{\max}:=\max_{i,j}|B_{ij}|$, and define
\begin{align}
 \ell_\theta(p,W)
 &:=\langle a,p\rangle+\frac12\langle B,W\rangle,
 &
 \|\theta\|_*
 &:=\max\{\|a\|_\infty,\tfrac12\|B\|_{\max}\},
 \label{eq:probe-energy}\\
 \gamma_\theta(G)
 &:=\min_{z\in\cS_q(G)}\ell_\theta(z),
 &
 \eta_q(G,G')
 &:=\sup_{\|\theta\|_*\leq1}
   |\gamma_\theta(G)-\gamma_\theta(G')|.
 \label{eq:scalar-probe-distance}
\end{align}
Support-function duality gives $\eta_q=\eta_{q,1}$.  We similarly view
$\eta_{q,2}$ through all scalar energies in the Euclidean dual unit ball,
and abbreviate $\delta_q:=\delta_{q,1}$ and $\kappa_q:=\kappa_{q,1}$.
Thus $\eta$ compares scalar energies, $\delta$ compares attainable-statistic
sets, and $\kappa$ measures the information lost by scalar optimization.

\begin{theorem}[Task-value size transfer]
\label{thm:task-value-size-transfer}
For all finite $D$-bounded graphs $G,G'$, every $q$, and $s\in\{1,2\}$,
\begin{equation}
 \eta_{q,s}(G,G')
 \leq \delta_{q,s}(G,G')
 \leq \eta_{q,s}(G,G')+\kappa_{q,s}(G)+\kappa_{q,s}(G').
 \label{eq:probe-quotient-sandwich}
\end{equation}
Consequently, if $F:\mathsf Y_{D,q}\to\mathbb R$ is $L_s$-Lipschitz
under $d_{q,s}$, then
\begin{equation}
 |\min_{z\in\cS_q(G)}F(z)-\min_{z'\in\cS_q(G')}F(z')|
 \leq L_s\bigl(\eta_{q,s}(G,G')+\kappa_{q,s}(G)+\kappa_{q,s}(G')\bigr).
 \label{eq:lipschitz-task-transfer}
\end{equation}
The same holds for maxima.  If $F$ is affine, the minimum and maximum
bounds hold with only $L_s\eta_{q,s}(G,G')$ on the right.  More explicitly, suppose
deleting at most $\varepsilon_G|V(G)|$ and
$\varepsilon_{G'}|V(G')|$ edges leaves components of orders at most
$K_G$ and $K_{G'}$, respectively.  For $s=1$, the rightmost quantity in
\eqref{eq:probe-quotient-sandwich} is at most
\begin{equation}
 \eta_q(G,G')+4(\varepsilon_G+\varepsilon_{G'})
 +2(1+D)r_q\bigl(
   K_G/|V(G)|+K_{G'}/|V(G')|
 \bigr).
 \label{eq:decomposable-size-transfer}
\end{equation}
For $s=2$, the $\ell^2$ analogue is the corresponding row of
Table~\ref{tab:transfer-regimes}. Its exact constant appears in
\eqref{eq:hyperfinite-convexification-l2}.
\end{theorem}

Some simplifications are possible under additional constraints: for a task reading only $k$ quotient coordinates, $r_q$ in
\eqref{eq:decomposable-size-transfer} improves to $k$
(Proposition~\ref{prop:projected-convexification}, Appendix~\ref{app:deferred-task}).
For affine tasks the $\kappa$ terms drop out of the transfer bound.
Applying this to the examples in
Section~\ref{sec:examples} we have
explicit constants on \emph{every} pair of graphs:

\begin{corollary}[Transfer of coloring, cut, and independence values]
\label{cor:opening-task-transfer}
Let $G,G'$ be finite $D$-bounded graphs, and let $J$ be a fixed nonempty
loopless graph on $h$ vertices.  Then, with $\eta_q\leq\delta_q$ by
\eqref{eq:probe-quotient-sandwich},
\begin{align}
 |e_q(G)-e_q(G')|
 &\leq\tfrac12\eta_q(G,G'),
 &
 |\operatorname{defect}_J(G)-\operatorname{defect}_J(G')|
 &\leq\tfrac12\eta_h(G,G'),
 \label{eq:color-defect-transfer}\\
 \Bigl|
  \tfrac{\operatorname{MaxCut}(G)}{|V(G)|}
  -\tfrac{\operatorname{MaxCut}(G')}{|V(G')|}
 \Bigr|
 &\leq\tfrac12\eta_2(G,G'),
 &
 \Bigl|
  \tfrac{\alpha(G)}{|V(G)|}
  -\tfrac{\alpha(G')}{|V(G')|}
 \Bigr|
 &\leq\eta_2(G,G').
 \label{eq:independence-transfer}
\end{align}
\end{corollary}

In particular, if $G$ is $q$-colorable, then deleting at most
$\frac12\eta_q(G,G')|V(G')|$ edges makes $G'$ $q$-colorable. The
compact-spin analogue is Corollary~\ref{cor:compact-spin-transfer}; proofs
are in Appendices~\ref{app:proof-task-value-transfer}
and~\ref{app:proof-compact-spin-transfer}.

\textbf{Rounding in neural combinatorial optimization.}
The expected quotient of any randomized assignment---e.g.\ the product
measure produced by a relaxation-based loss
\citep{karalias2020erdos,schuetz2022combinatorial,vanderbush2026neural}---lies in
$\operatorname{conv}\cS_q(G)$.  For an \emph{affine} loss the probabilistic
method rounds it without error on every graph. Theorem~\ref{thm:hyperfinite-convexification} extends this to rounding non-affine Lipschitz objectives, with an error controlled by the convexification bound.

\begin{corollary}[Rounding fractional quotient targets; informal]
\label{cor:rounding}
Under the decomposition hypothesis of
Theorem~\ref{thm:hyperfinite-convexification}, every fractional target
$z^\ast\in\operatorname{conv}\cS_q(G)$ is realized by a single
assignment up to $d_q$-error $4\varepsilon+2(1+D)r_qK/n$.  Hence every
Lipschitz task value follows suit, and on hyperfinite classes randomized and
single-assignment optima coincide asymptotically
(formal: Corollary~\ref{cor:rounding-formal}, Appendix~\ref{app:formal-size-transfer}).
\end{corollary}

\subsection{A quotient-aware graph-learning guarantee}
\label{sec:quotient-aware}

The next corollary spells out what an encoder and readout must supply
for training at bounded order and testing at any order.  Fix $J\geq1$ and a finite list of tasks
$T_j(G)=\operatorname{opt}_{z\in\cS_{q_j}(G)}F_j(z)$, $j\in[J]$, each
$F_j$ being $L_j$-Lipschitz and each optimum a minimum or maximum.  For
tuples $R=(K_1,\ldots,K_J)$, $R'=(K'_1,\ldots,K'_J)$ of compact sets,
write
$R(G):=(\cS_{q_1}(G),\ldots,\cS_{q_J}(G))$,
$d_R(R,R'):=\max_{j\in[J]}d_{\rm H}(K_j,K'_j)$, and
$L_*:=\max_{j\in[J]}L_j$.

\begin{corollary}[Deterministic size transfer]
\label{cor:finite-bank-size-generalization}
For every $r>0$ there is a finite training dataset $\mathcal B_r$ of finite
$D$-bounded graphs, of orders bounded by a number $n_{\max}(D,\mathbf q,r)$
independent of the test-graph order ($\mathbf q=(q_1,\ldots,q_J)$), such
that every finite $D$-bounded graph $G$ satisfies
$d_R(R(G),R(B))\leq r$ for some $B\in\mathcal B_r$.

Suppose an encoder produces
$\widehat R(G)\in\prod_{j=1}^J\cK(\mathsf Y_{D,q_j})$, with
$d_R(\widehat R(G),R(G))\leq\zeta$ for every graph, and a single readout
$f:\prod_{j=1}^J\cK(\mathsf Y_{D,q_j})\to\mathbb R^J$---with the same
parameters at every graph size---is $\Lambda$-Lipschitz for some
$\Lambda\geq0$, meaning that
$\|f(x)-f(y)\|_\infty\leq\Lambda d_R(x,y)$.  If
$\max_{B\in\mathcal B_r}
 \|f(\widehat R(B))-(T_1(B),\ldots,T_J(B))\|_\infty\leq\tau$,
then on every finite $D$-bounded graph $G$,
\begin{equation}
 \|f(\widehat R(G))-(T_1(G),\ldots,T_J(G))\|_\infty
 \leq \tau+(\Lambda+L_*)r+2\Lambda\zeta.
 \label{eq:finite-bank-generalization-bound}
\end{equation}
\end{corollary}

Training error on a finite, bounded-order dataset thus propagates to every
$D$-bounded test graph, with slack from the cover radius and encoder
accuracy (proof: Appendix~\ref{app:proof-task-value-transfer}). It remains to exhibit natural encoders that satisfy the above hypothesis.

\textbf{Training dataset construction.} Compactness alone makes the training
set existential. Furthermore, an $(\varepsilon,K(\varepsilon))$-decomposition
makes the construction finite, with $n_{\max}$ as in
\eqref{eq:explicit-nmax} (Proposition~\ref{prop:explicit-cover}). In particular, for the
class $\mathcal T_{D,t}$ of $D$-bounded graphs with treewidth at most $t$,
Appendix~\ref{app:bounded_treewidth}
gives the construction in full.

\textbf{Raw-graph realization.} We use a message-passing neural network (MPNN) with
random node initialization (RNI)
\citep{abboud2021surprising,sato2021random} to break
anonymous symmetry. However, it can also supply the consistent random ranks used by a
local partition oracle.  The next theorem compiles that observation into a
single network that receives only the original graph.

\begin{theorem}[Raw-graph RNI probe encoder]
\label{thm:raw-rni-encoder}
Fix a degree bound $D\geq2$, a treewidth bound $h\geq1$, a state number $q$, a finite
probe bank $\Theta\subseteq\{\theta:\|\theta\|_*\leq1\}$, and
$\alpha\in(0,1/2)$ and $\xi,\delta\in(0,1)$.  There is a fixed-weight RNI-MPNN
$\Psi=(\Psi_\theta)_{\theta\in\Theta}$, run
directly on the raw graph with vertex-mean readout, such that every graph
$G$ of maximum degree at most $D$ and treewidth at most $h$ satisfies
\begin{equation}
 \Pr\!\left[
  \max_{\theta\in\Theta}
  |\Psi_\theta(G)-\gamma_\theta(G)|\leq2\alpha+\xi
 \right]\geq1-\delta.
 \label{eq:raw-rni-probe-error}
\end{equation}
Its parameters are independent of $|V(G)|$.  More explicitly, one may take
an internal component scale and query budget
\begin{equation}
 K_\alpha=O\!\left(
  \frac{D^5h^{O(h)}\log(D/\alpha)}{\alpha^3}\right),
 \qquad
 Q_\alpha=O\!\left(DK_\alpha^{4h+7}\right),
 \label{eq:partition-oracle-complexity}
\end{equation}
depth $O(Q_\alpha+K_\alpha)$, and finite width depending only on the fixed
parameters above.  Note that this is a fixed-weight representation theorem. An optimization-level guarantee is still open.
\end{theorem}

The proof compiles the partition oracle of \citet{edelman2011partitioning}.
Beyond bounded treewidth, the next result exhibits a general obstruction on
expanders, whose treewidth grows.

\begin{theorem}[RNI depth lower bound for coloring defect]
\label{thm:rni-depth-lower}
There are fixed constants $D\geq3$ and $c_0,c_1>0$ and infinitely many
component orders $K$ with pairs $(A_{K,L},H_{K,L})$, for every $L\geq1$,
of $D$-regular graphs on the same number of vertices whose components have
order at most $2K$, such that
\begin{equation}
 e_2(A_{K,L})=0,
 \qquad e_2(H_{K,L})\geq c_0.
 \label{eq:rni-depth-separation}
\end{equation}
Let one $t$-layer local MPNN use iid node initialization, final node states
in $[-M,M]^w$, vertex-mean pooling, and a readout that is
$\Lambda$-Lipschitz for $\|\cdot\|_2$, with
finite $(w,M,\Lambda)$ independent of $L$.  If its expected absolute error
on both graph families is at most some $\eta<c_0/2$ for every $L$, then
$t\geq c_1\log K$.  The statement excludes global attention, virtual nodes,
positional IDs, and size-dependent state or readout bounds.
\end{theorem}
Their proofs appear in Appendices~\ref{app:proof-raw-rni-encoder}
and~\ref{app:proof-rni-depth-lower}; the compilation statement is in
Appendix~\ref{app:formal-size-transfer}.

\noindent\textbf{Empirical results.}
{
We evaluate exact energy features and trained neural baselines on a dataset
with maximum degree $3$ and treewidth $2$.  The energy probes are computed by
family-specific dynamic programs, so these experiments test the energy
representation rather than implement the RNI-MPNN encoder of
Theorem~\ref{thm:raw-rni-encoder}.  The dataset contains $72$ training graphs
with $n\in[8,30]$ and $66$ test graphs with $n\in[120,360]$.  Below, we report
test mean absolute error (MAE); learned-model entries are rounded means over
seeds $0,1,2$:
}
\par\smallskip
{\centering
\begin{tabular}{lccc}
\toprule
representation or model & $e_2$ & $\operatorname{MaxCut}/n$ & $\alpha/n$\\
\midrule
GPS--RWSE & $.0046$ & $.0085$ & $.0039$\\
spectral--motif ridge & {$.0092$} & {$.0092$} & $.0060$\\
$45$ exact generic energies + ridge & $.0005$ & $.0005$ & $.0002$\\
\bottomrule
\end{tabular}
\par}
\smallskip
The table separates representation quality from end-to-end learning.  The
$45$ exact probes are computed by family-specific dynamic programs and exclude
the energies that directly define the targets.  Their low MAE therefore shows
that generic energy coordinates retain the relevant task information, not that
we have learned these coordinates from raw graphs.  Among the learned models,
GPS--RWSE performs best and improves on the spectral--motif ridge (a handcrafted baseline using degree, triangle, and normalized-adjacency moment features) on all three
tasks, although the improvement for MAX-CUT is modest.  GPS--RWSE is trained only on graphs
with $n\leq30$ and evaluated on graphs with $n\in[120,360]$, so it demonstrates
size generalization to instances $4$--$12$ times larger than those seen during
training.

\noindent\textbf{Statistical generalization.}
{
The preceding guarantees are deterministic.  For a fixed bank of $P$ exact
probes, Proposition~\ref{prop:linear-generalization} gives, uniformly over
$\ell^1$-bounded linear readouts from $m$ iid $D$-bounded graphs labeled in
$[0,B]$, an absolute-risk gap at confidence $1-\alpha$ of
$O((W_1(1+D)+B)\sqrt{(\log P+\log(1/\alpha))/m})$.  Thus the rate is
$m^{-1/2}$, logarithmic in $P$, and independent of graph order.  At fixed
$D,B,\Lambda,P,\alpha$, Proposition~\ref{prop:probe-generalization}
gives $\widetilde O(m^{-1/(P+2)})$ for arbitrary $[0,B]$-valued,
$\Lambda$-Lipschitz readouts, equivalently
$\widetilde O(\epsilon^{-(P+2)})$ samples for gap $\epsilon$: the covering
cost of a $P$-dimensional cube, but still no graph-order dependence.  The
full-state analogue, Theorem~\ref{thm:generalization}, pays
$2^{(3(1+D)/\rho)^{q+q^2}}$ for one state size $q$.  Encoder and approximation
errors enter separately through the transfer bounds.
}

\section{Discussion}
\label{sec:discussion}
In this work, we show that the quotient set is the smallest representation that captures all finite-state pairwise tasks. On $(\varepsilon,K)$-decomposable graphs, scalar energy features approximate this representation at the rates of Theorem~\ref{thm:hyperfinite-convexification}, and these rates are essentially tight for constrained tasks. We further give representation-level guarantees: our encoder realizes the required information through message passing, while the depth lower bound identifies a limitation of bounded-state local architectures. Affine objectives continue to transfer on non-hyperfinite random-regular graphs, although whether gradient-based optimization reliably finds the corresponding solutions remains open. More broadly, important directions for further investigation are whether these results extend to richer graph limits such as graphops, and whether bounded-state energy hierarchies are strict and their probes can be learned efficiently.

\section*{Reproducibility statement}
The assumptions and normalization conventions used by the theoretical results
are stated with the definitions and theorem statements in
Sections~\ref{sec:attainable-regions}--\ref{sec:size-transfer}; formal versions
of the compressed size-transfer statements appear in
Appendix~\ref{app:formal-size-transfer}, and complete proofs of all deferred
claims are organized theorem-by-theorem in Appendix~\ref{app:routine-proofs}.
Appendices~\ref{app:experiment-details} and~\ref{app:trained-experiments}
document the procedurally generated graph families, data split, probe
construction, exact-routine safeguards, baseline features, neural architecture,
random seeds, optimization settings, and evaluation protocols for the
size-transfer comparison in Section~\ref{sec:size-transfer}.
We will release the code.

\newpage
\section*{Acknowledgments}
This research was developed with funding from the Defense Advanced Research
Projects Agency (DARPA) under agreements no. HR0011-25-3-0205 and
HR0011262E027. The views, opinions, and/or findings expressed are those of
the authors and should not be interpreted as representing the official views
or policies of the Department of Defense or the U.S. Government.
MW acknowledges partial support from an Alfred P. Sloan Fellowship in
Mathematics and the AI2050 program at Schmidt Sciences (Grant G-25-69786).

\section*{Use of generative AI}
The authors developed the research directions, identified connections between
sparse graph limits, the partition convergence of Borgs et al.\ and hyperfinite
graphs, and proposed graph neural network encoders with random node
initialization. Generative AI suggested convexification, introduced the
Shapley--Folkman literature, and developed the statistical generalization
theorem under the authors' guidance. It also assisted with proofs, experiments,
code, figures, literature review and writing. The authors checked the
mathematical statements and proofs, verified references and synthetic-data
properties, reviewed the code and exposition, and take responsibility for the
final work.

\bibliographystyle{plainnat}
\bibliography{references}

\clearpage
\appendix

\section{Affine vs nonlinear task: A worked translation to coloring}
\label{app:example_intro}

In this section, we detailed an application of the theory to a particular pairwise graph task: graph coloring.
The quotient formalism separates two coloring questions that are easy to
conflate: minimizing the number of monochromatic edges, and doing so while
controlling the color-class proportions.  The first objective is affine and
transfers on every graph.  The second is nonlinear and exposes the rounding
error created by convexification.  We translate the definitions and bounds
for both objectives.

\paragraph{Proper coloring as an affine task.}
Fix a palette $[q]$ and an assignment $c:V(G)\to[q]$.  Its quotient
coordinate $p_i$ is the fraction of vertices receiving color $i$, while
$W_{ii}/2$ is the number of monochromatic edges of color $i$, normalized by
$n:=|V(G)|$.  The factor $1/2$ removes the two orientations of each edge.
Consequently,
\begin{equation}
 u_q(p,W):=\frac12\sum_{i=1}^q W_{ii},
 \qquad
 e_q(G)=\min_{(p,W)\in\cS_q(G)}u_q(p,W)
 \label{eq:coloring-translation-defect}
\end{equation}
is the minimum monochromatic-edge density.  It vanishes exactly when $G$ is
properly $q$-colorable and otherwise equals the minimum number of edge
deletions per vertex needed to make $G$ $q$-colorable.

Because $u_q$ is affine, convexifying the quotient set does not change its
minimum.  The corresponding probe is $\theta_{\rm col}=(0,I_q)$, whose dual
norm in \eqref{eq:probe-energy} is $1/2$.  Thus
Corollary~\ref{cor:opening-task-transfer} gives, for every pair of
$D$-bounded graphs,
\begin{equation}
 |e_q(G)-e_q(G')|\leq \frac12\eta_q(G,G').
 \label{eq:coloring-translation-affine}
\end{equation}
In particular, if $G$ is properly $q$-colorable, deleting at most
$\frac12\eta_q(G,G')|V(G')|$ edges makes $G'$ $q$-colorable.  This
conclusion needs no decomposition assumption.  It concerns repair distance,
not the Boolean property of being $q$-colorable: for example,
$e_2(C_{2m+1})=1/(2m+1)\to0$, although every odd cycle has chromatic number
three (Appendix~\ref{app:disjoint-unions-edits}).

\paragraph{What a balance requirement adds.}
Fix a target color proportion $a\in\Delta_q$ and a finite penalty
$\lambda\geq0$.  Define the penalized balanced-coloring defect by
\begin{equation}
 \operatorname{BalDef}_{q,a,\lambda}(G)
 :=\min_{(p,W)\in\cS_q(G)}
 \bigl(u_q(p,W)+\lambda\|p-a\|_1\bigr).
 \label{eq:coloring-translation-balanced}
\end{equation}
Unlike \eqref{eq:coloring-translation-defect}, this objective is nonlinear
in the quotient.  A four-vertex star makes the lost information explicit.
The two proper $2$-colorings of $K_{1,3}$ have proportions $(1/4,3/4)$ and
$(3/4,1/4)$.  Their midpoint has proportion $(1/2,1/2)$ and zero defect, so
it belongs to $\operatorname{conv}\cS_2(K_{1,3})$.  No single balanced
assignment is proper, however: every two--two assignment leaves at least one
of the three edges monochromatic.  For $a=(1/2,1/2)$, direct enumeration
therefore gives
\begin{equation}
 \operatorname{BalDef}_{2,a,\lambda}(K_{1,3})
 =\min\{\lambda/2,1/4\},
 \qquad
 \operatorname{BalDef}_{2,a,\lambda}(K_{1,3}\sqcup K_{1,3})=0.
 \label{eq:star-balanced-coloring}
\end{equation}
The second graph colors its two copies in opposite orientations.  Yet
self-union leaves every convexified quotient set unchanged by
\eqref{eq:expander-convex-equality}, so all scalar energy features agree on
the two graphs.  Equation~\eqref{eq:star-balanced-coloring} is a concrete
instance of the set-to-hull information measured by $\kappa_2$.

\paragraph{The coloring-specific transfer bound.}
The balanced objective reads only $q+1$ statistics.  To expose those
statistics without losing constants, define the linear projection
\begin{equation}
 \Pi_q(p,W):=(p,\tau),
 \qquad \tau:=\sum_{i=1}^qW_{ii}=2u_q(p,W).
 \label{eq:coloring-projection}
\end{equation}
This projection is nonexpansive from the quotient $\ell^1$ metric to
$\ell^1$ because
$\|p-p'\|_1+|\tau-\tau'|\leq
\|p-p'\|_1+\|W-W'\|_1$.
Let $\kappa_q^{\rm col}(X)$ be the Hausdorff distance between
$\Pi_q\cS_q(X)$ and its convex hull, and put
$M_\lambda:=\max\{\lambda,1/2\}$.  The function
$(p,\tau)\mapsto\tau/2+\lambda\|p-a\|_1$ is
$M_\lambda$-Lipschitz.  Theorem~\ref{thm:task-value-size-transfer} therefore
specializes to
\begin{equation}
 \left|
 \operatorname{BalDef}_{q,a,\lambda}(G)
 -\operatorname{BalDef}_{q,a,\lambda}(G')
 \right|
 \leq M_\lambda\bigl(
 \eta_q(G,G')+\kappa_q^{\rm col}(G)+\kappa_q^{\rm col}(G')
 \bigr).
 \label{eq:balanced-coloring-transfer}
\end{equation}

Suppose that deleting at most $\varepsilon_X|V(X)|$ edges from each
$X\in\{G,G'\}$ leaves components of order at most $K_X$, and write
$n_X:=|V(X)|$.  Proposition~\ref{prop:projected-convexification}, applied to
the $q+1$ outputs of $\Pi_q$, bounds each coloring-specific gap by
\begin{equation}
 \kappa_q^{\rm col}(X)
 \leq 4\varepsilon_X+
 \frac{2(1+D)(q+1)K_X}{n_X}.
 \label{eq:coloring-projected-gap}
\end{equation}
Substitution into \eqref{eq:balanced-coloring-transfer} yields the explicit
size-transfer bound
\begin{align}
 &\left|
 \operatorname{BalDef}_{q,a,\lambda}(G)
 -\operatorname{BalDef}_{q,a,\lambda}(G')
 \right| \notag\\
 &\quad\leq M_\lambda\!\left[
 \eta_q(G,G')+4(\varepsilon_G+\varepsilon_{G'})
 +2(1+D)(q+1)\!\left(\frac{K_G}{n_G}+\frac{K_{G'}}{n_{G'}}\right)
 \right].
 \label{eq:coloring-decomposition-transfer}
\end{align}
Thus the task-specific projection improves the generic $q^2$ factor to
$q+1$.  The same estimate gives a rounding statement: every fractional
balance--defect profile in $\operatorname{conv}\Pi_q\cS_q(X)$ lies within
the right-hand side of \eqref{eq:coloring-projected-gap} of one deterministic
coloring profile, and its objective increases by at most $M_\lambda$ times
that distance.

For comparison, the full-space Euclidean bound in
\eqref{eq:hyperfinite-convexification-l2} gives the dimension-free branch in
Table~\ref{tab:transfer-regimes}.  A valid Euclidean Lipschitz constant for
\eqref{eq:coloring-translation-balanced} is
$L_2:=\sqrt{q(\lambda^2+1/4)}$.  Define the Euclidean representation
remainder of a decomposable graph $X$ by
\begin{equation}
 \beta_{q,2}(X):=4\varepsilon_X+\sqrt{1+D^2}
 \min\!\left\{\sqrt{r_q}\frac{K_X}{n_X},
                    \sqrt{\frac{K_X}{n_X}}\right\}
 \label{eq:coloring-euclidean-gap}
\end{equation}
and obtain
\begin{equation}
 \left|
 \operatorname{BalDef}_{q,a,\lambda}(G)
 -\operatorname{BalDef}_{q,a,\lambda}(G')
 \right|
 \leq L_2\bigl(
 \eta_{q,2}(G,G')+\beta_{q,2}(G)+\beta_{q,2}(G')
 \bigr).
 \label{eq:coloring-euclidean-transfer}
\end{equation}
On a growing hyperfinite family,
these remainders vanish after first choosing the deletion fraction small and
then taking the graph order large.  Scalar energies therefore preserve
ordinary coloring defect on every graph, while they preserve this
balance-sensitive coloring objective asymptotically when the quotient set
approaches its convex hull.  Exact balance constraints are not covered by
the finite-penalty statement.

\section{Extended related work}
\label{app:extended-related}

Section~\ref{sec:related-work} compresses the delimitation of prior art to its
quantified claims; this appendix gives the full exposition, one limit
notion at a time, with example- and theorem-level deltas.

\paragraph{Partition convergence.}
For simple $D$-bounded graphs, our node and oriented-edge weights $(p,W)$
are exactly the $(x,X)$ coordinates of \citet{borgs2017ld}, and our
Hausdorff convergence of all $\cS_q(G_n)$ is their \emph{partition
convergence}, up to an equivalent finite-dimensional norm. Thus the topology
and its compact hyperspace packaging are prior work, not a new limit notion.
Their Examples~4 and~5 also separate partition and Benjamini--Schramm
convergence \citep{benjamini2001recurrence} in both directions: alternating one expander copy with two
preserves a local limit but destroys partition convergence, while
alternating unions of $4$- and $6$-cycles is partition convergent but not
locally convergent.

\paragraph{Normalized quotients and ground-state energies.}
The quotient calculus originates in the dense theory:
\citet{bclsv2012} prove that cut-metric convergence, Hausdorff convergence
of quotient sets, and \emph{microcanonical} ground-state and free-energy
convergence are equivalent for dense sequences (their Theorem~2.8), that
plain (unconstrained) energies are strictly weaker (their Corollary~3.6
and Examples~6.1 and~6.3---the dense ancestors of our
Corollary~\ref{cor:necessity}), and
that energies are Lipschitz in cut distance (their Theorem~4.6, the dense
ancestor of Corollary~\ref{cor:opening-task-transfer}); their fractional
quotient sets are closures of integer ones, not convex hulls, and convexity
plays no role in that regime.  For bounded degree,
\citet{borgs2013left} relate right and left convergence through free
energies of soft-core models, again without quotient geometry.
\citet[Definition~2.3]{borgs2018lp2} extend the dense equivalences to
sparse graphs with unbounded average degree under uniform upper regularity,
calling Hausdorff convergence of their
quotient sets \emph{quotient convergence}. For a nonempty simple graph their
edge coordinate is
\begin{equation}
 \beta_{ij}=\frac{W_{ij}}{\bar d(G)},
 \qquad \bar d(G):=\frac{2|E(G)|}{|V(G)|},
 \label{eq:bccz-normalization}
\end{equation}
whereas we retain $W_{ij}$.  Neither theory has content at bounded
degree, for two different reasons: the dense normalization divides edge
counts by $n^2$, so every $D$-bounded quotient collapses to zero edge
weights and each side of the dense equivalence holds vacuously there,
while a uniformly upper-regular sequence of simple graphs has unbounded
average degree \citep[Proposition~C.15]{borgs2019lp1}, so $D$-bounded
sequences sit outside the $L^p$ class altogether.  Under the vertex
normalization, which keeps bounded degree meaningful, the equivalence
acquires content---and in that form it is false, by the expander pair of
Proposition~\ref{prop:expander-separation}; only the reverse direction is
at issue, the forward direction being free of hypotheses (their
Theorem~3.3, or Theorem~\ref{thm:task-completeness} here).  Two of their examples are the nearest
precursors of this paper: their Example~3.3.5 shows unconstrained energies
insufficient even under upper regularity, and their Example~3.3.7 (after
\citealp{borgs2017ld}) observes, qualitatively, that quotients of growing
disjoint unions converge to the \emph{convex hull} of the component
quotients---Theorem~\ref{thm:hyperfinite-convexification} turns exactly
this observation into a universal finite-graph upper bound for arbitrary
$(\varepsilon,K)$-decompositions, with worst-case $q$- and $K/n$-dependences
matched separately. Their Theorem~3.3 proves that normalized quotient
convergence implies convergence of ordinary and microcanonical ground-state
energies, and their Theorem~2.15 makes metric convergence, quotient
convergence, and \emph{microcanonical} ground-state convergence equivalent
for uniformly upper-regular weighted sequences. On $D$-bounded sequences,
partition convergence forces $\bar d(G_n)$ to converge (the total $W$-mass is
$\bar d$); when the limit is positive the two quotient coordinates differ
only by a convergent scalar, so on such sequences the topologies essentially
agree.  The delta claimed here is therefore not the topology but the
\emph{energy family and its quantitative reach}: (i)
Corollary~\ref{cor:hyperfinite-scalar-completeness} characterizes partition
convergence on hyperfinite classes by \emph{unconstrained} unary--pairwise
energies alone, a strictly weaker family than the microcanonical one---and
Corollary~\ref{cor:necessity} shows by an expander alternation that this
weaker family cannot suffice in general, so the hyperfiniteness hypothesis is
doing real work; (ii)
Proposition~\ref{prop:lagrangian-bridge} bounds the penalized
microcanonical-versus-bounded-Lagrangian gap by the convexification error
$\kappa_q$, turning the relation between the two energy families into a
finite-graph inequality rather than a limit statement; and (iii) all of this
comes with the explicit finite-$n$ rate of
Theorem~\ref{thm:hyperfinite-convexification} and the matching lower bound of
Remark~\ref{thm:sharp-convexification}, neither of which appears in the
cited works.  The distinction in (i) matters for learning: a microcanonical
energy is not a feature map---a learner's energy-based features are
unconstrained optima---so the classical characterizations, correct as they
are, are not implementable as order-independent feature vectors, and the
question of when the implementable family suffices is not posed there.

\paragraph{Small-component decompositions and hyperfiniteness.}
Hyperfiniteness is standard terminology for bounded-degree graph families
that can be broken into uniformly small components after deleting few edges
\citep{schramm2008hyperfinite,newman2013hyperfinite,hatami2014localglobal}.
\citet[Theorem~10.5 and Corollary~10.6]{hatami2014localglobal} prove that a
growing, locally convergent hyperfinite bounded-degree sequence is
local--global convergent. Since $\cS_q$ is a continuous first-moment image of
the radius-one colored-neighborhood set, their theorem implies partition
convergence under those hypotheses. Our Theorem~\ref{thm:hyperfinite-convexification}
instead gives a finite-graph distance to $\operatorname{conv}\cS_q$, and
Corollary~\ref{cor:hyperfinite-scalar-completeness} characterizes partition
convergence on a hyperfinite sequence by convergence of all scalar
pairwise ground-state energies. This quantitative convexification and scalar
characterization are not statements of Theorem~10.5.

Our novelty audit covered the dense series and its bounded-degree and
$L^p$ successors above \citep{bclsv2012,borgs2013left,borgs2017ld,borgs2018lp2},
intrinsic partition representations by measures and square-graphons
\citep{kunszenti2019measures,dolezal2026categorical}, hyperfinite graphings
in combinatorial optimization \citep{lovasz2017hyperfinite}, the
local--global and action literature cited below, the generic
Shapley--Folkman theorem \citep{starr1969quasi}, the classical
Shapley--Folkman duality-gap estimates for separable nonconvex programs
\citep{aubin1976estimates,udell2016bounding}, of which
Proposition~\ref{prop:lagrangian-bridge} is a quotient-set instantiation,
and overlap-gap-property obstructions for local algorithms on sparse random
regular graphs and for fixed-depth graph neural networks on particular random
structures \citep{gamarnik2017limits,gamarnik2023barriers}, which block
\emph{optimization} in their stated random-instance regimes and are
complementary to the \emph{representation} ceiling certified here. Within that searched scope, we did not find a prior
finite-graph bound on
$d_{\rm H}(\cS_q(G),\operatorname{conv}\cS_q(G))$ from a bounded-component
edge decomposition, nor the resulting scalar-energy characterization without
an assumed local limit. This is a scoped absence claim, not an exhaustive
claim over all decomposable optimization.  On sharpness, the $K/n$ term is
matched at fixed $q$, the $r_q=\Theta(q^2)$ factor is matched on perfect
matchings, and the Euclidean rate is matched in both of its regimes
(Remark~\ref{thm:sharp-convexification}); hyperfiniteness is non-removable
(Corollary~\ref{cor:necessity}).  The optimal constant in the
$4\varepsilon$ term and the optimal number of finite probes remain open.
Ground-state coordinates are efficiently computable on bounded treewidth
(Proposition~\ref{prop:treewidth-encoder}) and generally hard beyond it.

\paragraph{Partition oracles, Baker's technique, and the convex hull.}
On hyperfinite classes, partition oracles
\citep{hassidim2009local,newman2013hyperfinite} give sublinear-query
access to an $(\varepsilon,K)$-decomposition, and every additive graph
parameter---MAX-CUT, minimum vertex cover, maximum independent set---is
then estimated to additive $\varepsilon n$ from the empirical distribution
of component isomorphism types; Baker's layering
\citep{baker1994approximation} and separator recursions
\citep{lipton1980applications} give classical polynomial-time approximation
schemes for many planar optimization problems.  Qualitatively, then, it is
classical that on hyperfinite graphs each task value is a local statistic
of component types, and Proposition~\ref{prop:treewidth-encoder} and
Proposition~\ref{prop:explicit-cover} re-derive that statement in quotient
language.  Theorem~\ref{thm:hyperfinite-convexification} says something
different in two respects.  First, it concerns the \emph{representation}
rather than any one task: if $N_K$ denotes the number of connected
$D$-bounded types of order at most $K$, the component-type histogram has
dimension $N_K\leq K2^{K(K-1)/2}$.  By contrast, the convex hull of the
quotient set lives in dimension $r_q=\Theta(q^2)$ and is read off, up to
error $2(1+D)\rho$, by the $P=|\mathcal N_\rho|$ scalar energies in any
finite $\rho$-net $\mathcal N_\rho$ of the dual-unit probe ball.  The theorem gives the universal compression-loss upper bound
$4\varepsilon+2(1+D)r_qK/n$ for \emph{all $1$-Lipschitz} $q$-state pairwise
tasks simultaneously; an $L$-Lipschitz task multiplies the bound by $L$.
Separate examples match the worst-case $K/n$ and $q^2$
dependences; the optimal $\varepsilon$ coefficient remains open.  This includes
the constrained ones (balanced cut, penalized prescribed-balance
coloring) that are not additive parameters and that the partition-oracle
estimators do not address.  Second, the object that suffices is the
family of \emph{unconstrained energies}, not the histogram; the classical
equivalences reviewed above required microcanonical energies for the
same conclusion.  For bounded treewidth, Theorem~\ref{thm:raw-rni-encoder}
compiles the explicit local oracle of \citet{edelman2011partitioning} into
a fixed-weight RNI-MPNN on the raw graph; this is a representational
compilation, not a faster partition oracle.  The matching lower bounds of
Remark~\ref{thm:sharp-convexification} and the expander separation of
Proposition~\ref{prop:expander-separation}
have no counterpart in the partition-oracle literature, whose guarantees
are one-sided.  Conversely, we claim no algorithmic improvement over
partition oracles: the encoder has the same query scale, and
Theorem~\ref{thm:rni-depth-lower} shows that $\Omega(\log K)$ local depth
can be necessary for bounded-state RNI networks with Lipschitz mean readout
that must work uniformly over disjoint-union repetitions.

\paragraph{Size transfer and graph-learning generalization.}
In the GNN literature, \emph{size transferability} compares one fixed model
on graphs or discretizations of different orders
\citep{le2023graphoptransfer}; the cited graphop theorem compares regular
discretizations of one common limiting operator.  By contrast,
\emph{size generalization} commonly
means learning on smaller graphs and predicting larger ones
\citep{yehudai2021size}.  These notions are distinct from statistical
empirical-to-population generalization, as studied for attributed DIDM and
graphop representations by \citet{rauchwerger2025attributed,amran2026graphop}.
Section~\ref{sec:size-transfer} is target-side: it bounds normalized optimum
values, and then gives a deterministic generalization certificate for a
separately quotient-stable predictor.  It neither proves stability of an
arbitrary GNN nor supplies an optimization algorithm.  A further relevant
line is learning-based combinatorial optimization, where relaxation losses
train networks to output fractional assignments whose expected objective is
then rounded \citep{karalias2020erdos,schuetz2022combinatorial}; the
normalized task values studied here are exactly the targets of such
pipelines, and Corollary~\ref{cor:rounding} states the corresponding
target-side rounding guarantee on decomposable graphs.

\textbf{Expressivity and symmetry breaking.}
Anonymous message passing is bounded by the $1$-WL test
\citep{xu2019powerful,morris2019weisfeiler}; random node initialization
and unique identifiers restore universality at bounded order
\citep{abboud2021surprising,sato2021random,loukas2020what}.  We use this
bounded-order universality to compile a local bounded-treewidth partition
oracle \citep{edelman2011partitioning} into a raw-graph RNI-MPNN
(Theorem~\ref{thm:raw-rni-encoder}); it does not by itself imply the
variable-size result.  General depth--width lower bounds for message passing
are known \citep{loukas2020what}; Theorem~\ref{thm:rni-depth-lower} gives a
task-specific $\Omega(\log K)$ obstruction for bounded-state RNI networks
with Lipschitz mean readout that must work uniformly over disjoint-union
repetitions.

Accordingly, the paper does not replace these richer topologies.  Its new
representation statements are restricted to finite-state pairwise task
values and, for the raw-graph neural realization, bounded-treewidth inputs.

\section{Extensions and omitted details}
\label{app:details}

This appendix collects the material the main text defers: worked quotient
computations, the comparison with richer limit topologies, effective
encoders, per-section deferred discussion, the geometry of the
convexification error, and the experimental protocol.  Each subsection
opens from the main-text passage that points to it.

\subsection{Quotient sets in four simple examples}\label{app:example_quotient_set}
\paragraph{A single edge.}
Assign states $1$ and $2$ to the two endpoints of $K_2$.  Then
$p_1=p_2=1/2$ and $W_{12}=W_{21}=1/2$: the two off-diagonal entries record
the two orientations of the same undirected edge.  This example also
explains the factor $1/2$ that appears in undirected edge energies.

\paragraph{Edgeless graphs.}
For the edgeless graph $I_n$, the quotient set is the denominator-$n$ grid of
points $(p,0)$ and therefore
\begin{equation}
 \cS_q(I_n)\longrightarrow\Delta_q\times\{0\}.
 \label{eq:isolates-limit}
\end{equation}
\paragraph{Cycles.}
For a binary assignment on $C_n$, if $j$ vertices receive state $1$ in $r$
cyclic runs, then $(p_1,W_{12})=(j/n,2r/n)$.  Hence
$\cS_2(C_n)$ projects to
$\{(j/n,2r/n):1\leq r\leq\min(j,n-j)\}\cup\{(0,0),(1,0)\}$ and converges to
$\{(p_1,W_{12}):W_{12}\leq2\min(p_1,1-p_1)\}$, as shown in
Figure~\ref{fig:quotient-construction}(d).

\paragraph{Perfect matching.}
If $M_n$ is a perfect matching on an even number $n$ of vertices, then
\begin{equation}
 \cS_q(M_n)\longrightarrow
 \{(p,W):p\in\Delta_q,\ W\geq0,\ W=W^\top,\ W\mathbf1=p\},
 \label{eq:matching-limit}
\end{equation}
where $\mathbf1$ is the all-ones vector; rational symmetric matrices with the
displayed row sums are realized by choosing colored edge-type counts.  The
two examples share limiting state proportions but not edge weights, so state
proportions alone cannot preserve a pairwise interaction.

\subsection{Topological comparison of quotient space}\label{app:quotient_in_local_global_action}

\paragraph{First-moment factor of local--global and action profiles.}
The quotient is deliberately coarser than a colored rooted-neighborhood law.
Let $O$ be a uniformly sampled vertex, let $L_1^{G,c}$ be the law of its
colored radius-one neighborhood, and let
\begin{equation}
 \cQ_{q,1}(G):=\{L_1^{G,c}:c:V(G)\to[q]\}
 \label{eq:radius-one-local-global}
\end{equation}
be the radius-one local--global coordinate.  A continuous linear map $T_q$
reads from such a law
\begin{equation}
 p_i=\Pr(c(O)=i),\qquad
 W_{ij}=\mathbb E\!\left[
  \mathbf1_{\{c(O)=i\}}\#\{u\sim O:c(u)=j\}
 \right].
 \label{eq:root-moment-map}
\end{equation}
Consequently,
\begin{equation}
 \cS_q(G)=T_q(\cQ_{q,1}(G)).
 \label{eq:S-factor-Q}
\end{equation}

The same map appears inside graphop action convergence.  Regard the
unnormalized adjacency matrix $A_G$ as an operator on the uniform vertex
probability space and put $v_i:=\mathbf1_{\{c=i\}}$.  The action observation
records the joint law of
$(v_1,\ldots,v_q,A_Gv_1,\ldots,A_Gv_q)$, while the quotient keeps only
\begin{equation}
 p_i=\mathbb E[v_i(O)],
 \qquad
 W_{ij}=\mathbb E[v_i(O)(A_Gv_j)(O)].
 \label{eq:action-to-quotient}
\end{equation}
Equation~\eqref{eq:action-to-quotient} is a pointwise moment identity from a
hard one-hot observation to one quotient; it does not assert continuity of
the hard slice for convergent action profiles.

\subsection{Exact computations on bounded treewidth and explicit training sets}\label{app:bounded_treewidth}
\begin{proposition}[Exact quotient computation on bounded treewidth]
\label{prop:treewidth-encoder}
Fix $q\geq2$, $D$, $t$, and $\rho>0$.  Let $\mathcal T_{D,t}$ be the class of
$D$-bounded graphs of treewidth at most $t$, and let $\mathcal N_\rho$ be a
finite $\rho$-net of $\{\theta:\|\theta\|_*\leq1\}$ in the norm
$\|\cdot\|_*$.  Given $G\in\mathcal T_{D,t}$ on $n$ vertices together
with a width-$t$ tree decomposition, the full quotient set $\cS_q(G)$
can be listed exactly
by dynamic programming.  Consequently, on
$\mathcal T_{D,t}$: (i) the encoder $\widehat R=R$ of
Corollary~\ref{cor:finite-bank-size-generalization} is realizable with
$\zeta=0$ in polynomial time for fixed $(q,t,D)$; (ii) every probe energy
$\gamma_\theta(G)$ in \eqref{eq:scalar-probe-distance} is computable in the
same time; and (iii)
for all
$G,G'\in\mathcal T_{D,t}$ with orders $n,n'$,
\begin{equation}
 \delta_q(G,G')\;\leq\;
 \max_{\theta\in\mathcal N_\rho}|\gamma_\theta(G)-\gamma_\theta(G')|
 +2(1+D)\rho
 +c_{q,D,t}\Bigl(n^{-1/2}+n'^{-1/2}\Bigr),
 \label{eq:effective-certificate}
\end{equation}
where $c_{q,D,t}:=8\sqrt{(1{+}D)r_q(t{+}1)D}+2(1{+}D)r_q$.
\end{proposition}

\paragraph{Explicit training sets on decomposable classes.}
Corollary~\ref{cor:finite-bank-size-generalization} obtains the training
set $\mathcal B_r$ from compactness, which bounds neither its size nor the
order $n_{\max}$ of its members.  On classes with an explicit
decomposition both become explicit, because
Theorem~\ref{thm:hyperfinite-convexification} reduces every quotient set to
a convex combination of the finitely many quotient sets of connected
graphs of order at most $K$.

\begin{proposition}[Explicit training sets]
\label{prop:explicit-cover}
Let $\mathcal C$ be a class of $D$-bounded graphs such that, for every
$\varepsilon>0$, deleting at most $\varepsilon n$ edges of any
$G\in\mathcal C$ leaves components of order at most $K(\varepsilon)$.
Fix $q$ and $r>0$, put $\varepsilon:=r/8$, $K:=K(\varepsilon)$, let
$\mathcal T_K$ be the set of isomorphism types of connected $D$-bounded
graphs of order at most $K$, $N_K:=|\mathcal T_K|\leq K\,2^{K(K-1)/2}$,
and
\begin{equation}
 n_0:=\Bigl\lceil\frac{8(1+D)K\,(r_q+N_K)}{r}\Bigr\rceil,\qquad
 n_{\max}:=n_0+N_KK.
 \label{eq:explicit-nmax}
\end{equation}
Let $\mathcal B_r$ be the (finite) set of all disjoint unions of graphs
from $\mathcal T_K$ with at most $n_{\max}$ vertices.  Then every
$G\in\mathcal C$ satisfies $d_{\rm H}(\cS_q(G),\cS_q(B))\leq r$ for some
$B\in\mathcal B_r$.  For a task tuple with state spaces $q_1,\ldots,q_s$
the same $\mathcal B_r$ works with $r_q$ replaced by $\max_jr_{q_j}$, so
Corollary~\ref{cor:finite-bank-size-generalization} holds on $\mathcal C$
with the explicit $n_{\max}$ of \eqref{eq:explicit-nmax}.
\end{proposition}

Every member of $\mathcal B_r$ has treewidth at most $K-1$, so its quotient
set is computable by Proposition~\ref{prop:treewidth-encoder} with this fixed
bound.  For the specialization $\mathcal C=\mathcal T_{D,t}$, restrict
$\mathcal T_K$ to the types of treewidth at most $t$; the proof is unchanged,
and then $\mathcal B_r\subset\mathcal T_{D,t}$.
The constant $N_K$ is crude---for $D=2$ the connected graphs of order at
most $K$ are paths and cycles, so $N_K\leq2K$---and the proof (Appendix~\ref{app:proof-explicit-cover})
loses a factor $N_K$ only in rounding the component-type proportions to a
common denominator.

\subsection{Formal statements for Section~\ref{sec:size-transfer}}
\label{app:formal-size-transfer}

The informal statements of Section~\ref{sec:size-transfer} compress the
following.

\begin{corollary}[Rounding fractional quotient targets; formal]
\label{cor:rounding-formal}
Suppose deleting at most $\varepsilon n$ edges from a $D$-bounded $G$ leaves
components of order at most $K$, and let $z^\ast\in\operatorname{conv}\cS_q(G)$
be any fractional target.  Then there is an assignment
$c:V(G)\to[q]$ with
\begin{equation}
 d_q\bigl(G/c,\;z^\ast\bigr)
 \;\leq\;4\varepsilon+\frac{2(1+D)r_qK}{n},
 \label{eq:rounding}
\end{equation}
and hence $F(G/c)\leq F(z^\ast)+L\bigl(4\varepsilon+2(1+D)r_qK/n\bigr)$ for
every $L$-Lipschitz $F:\mathsf Y_{D,q}\to\mathbb R$.  On hyperfinite classes
the attainable-in-expectation and attainable-by-one-assignment task values
therefore coincide asymptotically.
\end{corollary}

\begin{proof}
The distance claim is Theorem~\ref{thm:hyperfinite-convexification}, applied
to $z^\ast\in\operatorname{conv}\cS_q(G)$; the task bound is the
$L$-Lipschitz inequality for $F$.
\end{proof}

The raw-graph theorem uses the following compilation step.  It is stated
separately because bounded-order universality alone does not justify a
single network on graphs of arbitrary order.

\begin{lemma}[Compiling a bounded local algorithm]
\label{lem:local-rni-compilation}
Fix $D,Q,P$ and $0<B<\infty$.  Let $\mathcal A$ be a comparison-based rooted
local algorithm that, starting at its root, makes at most $Q$ adjacency-list
queries in a $D$-bounded graph and queries only vertices reached from that
root.  Suppose it uses two independent iid $U[0,1]$ ranks per vertex---one
for its randomized decisions and one to break every BFS or enumeration tie,
including loops over discovered vertex sets---and
returns an isomorphism-equivariant vector $a_G(v)\in[-B,B]^P$.  For every
$\xi,\zeta>0$ there is a finite-width
RNI-MPNN of depth $O(Q)$, with parameters independent of graph order, whose
root outputs $\widehat a_G(v)$ satisfy, under a coupling with the same ranks,
\begin{equation}
 \Pr\!\left[
  \left\|\frac1n\sum_v\widehat a_G(v)-
               \frac1n\sum_v a_G(v)\right\|_\infty\leq\xi
 \right]\geq1-\zeta
 \label{eq:local-rni-compilation}
\end{equation}
for every finite input graph $G$.
\end{lemma}

\subsection{Architecture behind the raw-graph encoder}

The construction in Theorem~\ref{thm:raw-rni-encoder} starts from iid
ranks, which seed a consistent local partition oracle; at a root $v$, the network
reconstructs its oracle part $C(v)$ and returns
$(\gamma_\theta(C(v)))_{\theta\in\Theta}$.  Vertex-mean pooling is not
learned: for the partitioned graph $H$ it obeys the exact identity
\begin{equation}
 \frac1n\sum_{v\in V(G)}\gamma_\theta(C(v))
 =\sum_{C\in H}\frac{|C|}{n}\gamma_\theta(C)
 =\gamma_\theta(H).
 \label{eq:component-probe-mean}
\end{equation}
The deleted edges then cost at most $2\alpha$.  Independent channels and a
coordinatewise median amplify the oracle's constant success probability to
$1-\delta$.  Define the exact vector
$\gamma_\Theta(X):=(\gamma_\theta(X))_{\theta\in\Theta}$.  Every
$1$-Lipschitz head
$\psi:\ell^\infty(\Theta)\to\mathbb R$ obeys
$|\psi(\gamma_\Theta(G))-\psi(\gamma_\Theta(G'))|\leq\delta_q(G,G')$.
For any fixed pair, the theorem and a union bound add the corresponding
approximation errors with high probability; no uniform-over-graphs claim is
made.

\subsection{Deferred material for Section~\ref{sec:attainable-regions}}
\label{app:deferred-attainable}

\paragraph{The partition limit space.}
`Hyperspaces' are spaces of subsets of another space; $\cK(Y)$ denotes the
hyperspace of nonempty compact subsets of a compact metric space $Y$.
The \emph{finite-graph partition state, ambient product, and
finite-realizable limit space} of Section~\ref{sec:attainable-regions}
are, respectively,
\begin{align}
 \Theta_D^{\mathrm{part}}(G)
 :=\bigl(\cS_q(G)\bigr)_{q\geq1},\quad
 \mathfrak A_D^{\mathrm{part}}
 :=\prod_{q\geq1}\cK(\mathsf Y_{D,q}),
 \quad
 \mathfrak X_D^{\mathrm{part}}
 :=\overline{\{\Theta_D^{\mathrm{part}}(G):\Delta(G)\leq D\}}.
 \label{eq:partition-limit-space}
\end{align}
For $x\in\mathfrak X_D^{\mathrm{part}}$, write $\cS_q(x)$ for its $q$th
coordinate.  The metric
\begin{equation}
 d_D^{\mathrm{part}}(x,y)
 :=\sum_{q\geq1}2^{-q}
 \min\{1,d_{\rm H}(\cS_q(x),\cS_q(y))\}
 \label{eq:partition-metric}
\end{equation}
induces the product topology (Figure~\ref{fig:quotient-construction}(d));
Appendix~\ref{app:quotient_in_local_global_action} locates quotients
inside the richer local--global \citep{hatami2014localglobal} and action
\citep{backhausz2022action} topologies.  The packaging exists so that
limits never have to be assumed (proof in Appendix~\ref{app:proof-compactness}).

\begin{theorem}[Compactness]
\label{thm:compactness}
The ambient space $\mathfrak A_D^{\mathrm{part}}$ and the finite-realizable
space $\mathfrak X_D^{\mathrm{part}}$ are compact metrizable.  Every sequence
of finite $D$-bounded graphs has a partition-convergent subsequence.
\end{theorem}

Every construction in this paper can therefore extract convergent
subsequences with no structural hypothesis on the graphs.

\paragraph{One coordinate per task.}
Permutation invariance is restored at the set level: quotient sets are
unions of permutation orbits (worked examples in
Appendix~\ref{app:example_quotient_set}).
A fixed finite-state task needs only one quotient coordinate.  If $Q\geq q$
and $s:[Q]\to[q]$ is a surjection, aggregating node and edge weights over the
fibers of $s$ maps $\cS_Q(G)$ onto $\cS_q(G)$.  The countable family is kept
only to cover all finite state spaces at once.  Color names add no
information: each quotient set is invariant under simultaneously permuting
$p$ and the rows and columns of $W$, and because permutations act by
$\ell^1$-isometries and quotient sets are unions of permutation orbits, the
Hausdorff distance computed from $d_q$ coincides with the distance between
the label-free orbit quotients.

\subsection{Sharpness and deferred material for
Section~\ref{sec:task-calculus}}
\label{app:deferred-task}
\label{sec:sharpness}

\paragraph{Formal statements behind Remark~\ref{thm:sharp-convexification}
and Proposition~\ref{prop:expander-separation}.}
Let $M_{2m}=mK_2$ be a perfect matching, put
$N_q:=\binom{q+1}{2}$ and
$s_*:=\min\{m,\lfloor N_q/2\rfloor\}$, and define
\begin{equation}
 R_{q,m}:=\frac{2s_*(N_q-s_*)}{N_qm}.
 \label{eq:type-grid-radius}
\end{equation}
For every $q\geq2$ and $m\geq1$,
\begin{equation}
 R_{q,m}\leq\kappa_q(M_{2m})\leq2R_{q,m},
 \qquad
 \kappa_q(M_{2m})
 =\Theta\!\left(\min\{1,q^2/m\}\right).
 \label{eq:matching-l1-sharpness}
\end{equation}
If $X$ is a connected $D$-regular graph on $K$ vertices with edge
expansion at least $h_0>0$, meaning
$|\partial S|\geq h_0\min\{|S|,|V(X)\setminus S|\}$ for every
$S\subseteq V(X)$, where $\partial S$ is the set of edges with exactly one
endpoint in $S$, and $G=mX$, then, for $n=mK$,
\begin{equation}
 \kappa_2(G)\geq
 \min\{\tfrac12,\tfrac{h_0}{2}\}\frac Kn.
 \label{eq:component-lower-bound}
\end{equation}
Finally, every $D$-bounded $n$-vertex graph for which deleting at most
$\varepsilon n$ edges leaves components of order at most $K$ satisfies
\begin{equation}
 \kappa_{q,2}(G)
 \leq4\varepsilon+\sqrt{1+D^2}\,
 \min\!\left\{\sqrt{r_q}\frac Kn,\sqrt{\frac Kn}\right\}.
 \label{eq:hyperfinite-convexification-l2}
\end{equation}
Moreover,
$\kappa_{q,2}(M_{2m})=\Theta(\min\{m^{-1/2},\sqrt{N_q}/m\})$.

For the separation, let $G_n$ be a finite graph and
$H_n:=G_n\sqcup G_n$.  For every $q$,
\begin{equation}
 \operatorname{conv}\cS_q(H_n)=\operatorname{conv}\cS_q(G_n).
 \label{eq:expander-convex-equality}
\end{equation}
If $G_n$ has edge expansion at least $h>0$, meaning
$|\partial S|\geq h\min\{|S|,|V(G_n)\setminus S|\}$ for every vertex set
$S$, where $\partial S$ contains the edges with exactly one endpoint in
$S$, then the continuous balanced-cut value
\begin{equation}
 \Phi(G):=\min_{(p,W)\in\cS_2(G)}
 \left(|p_1-\tfrac12|+W_{12}\right)
 \label{eq:balanced-cut-query}
\end{equation}
satisfies $\Phi(H_n)=0$ and $\Phi(G_n)\geq\min\{1/2,h/2\}$.

\paragraph{The Shapley--Folkman mechanism.}
Put $r_q:=(q-1)+\frac{q(q+1)}2-1$.  For each fixed graph, this is a common
dimension bound after translating its component quotient sets: $p$ has fixed
sum one, $W$ is symmetric, and the total mass of $W$ is fixed on each
component.  In particular, $r_1=0$.  The proof of
Theorem~\ref{thm:hyperfinite-convexification} uses the following standard
weighted form of the Shapley--Folkman lemma \citep{starr1969quasi}: if
$A_1,\ldots,A_s\subset\mathbb R^m$ and $\lambda_j\ge0$, then every point of
$\sum_j\lambda_j\operatorname{conv}A_j$ can be written with at most $m$
summands still chosen from $\operatorname{conv}A_j$ and every other summand
chosen from $A_j$.

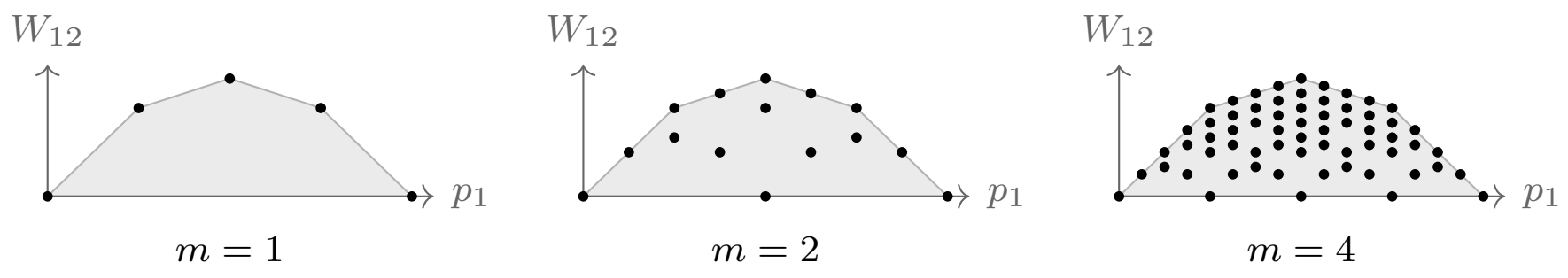
\begin{figure}[t]
\centering
\begin{tikzpicture}
\foreach \m/\xs/\pts in {%
 1/0.35/{(0,0),(0.25,0.75),(0.5,1),(0.75,0.75),(1,0)},%
 2/4.35/{(0,0),(0.125,0.375),(0.25,0.5),(0.25,0.75),(0.375,0.375),(0.375,0.875),(0.5,0),(0.5,0.75),(0.5,1),(0.625,0.375),(0.625,0.875),(0.75,0.5),(0.75,0.75),(0.875,0.375),(1,0)},%
 4/8.35/{(0,0),(0.0625,0.1875),(0.125,0.25),(0.125,0.375),(0.1875,0.1875),(0.1875,0.4375),(0.1875,0.5625),(0.25,0),(0.25,0.375),(0.25,0.5),(0.25,0.625),(0.25,0.75),(0.3125,0.1875),(0.3125,0.4375),(0.3125,0.5625),(0.3125,0.6875),(0.3125,0.8125),(0.375,0.25),(0.375,0.375),(0.375,0.625),(0.375,0.75),(0.375,0.875),(0.4375,0.1875),(0.4375,0.4375),(0.4375,0.5625),(0.4375,0.6875),(0.4375,0.8125),(0.4375,0.9375),(0.5,0),(0.5,0.375),(0.5,0.5),(0.5,0.625),(0.5,0.75),(0.5,0.875),(0.5,1),(0.5625,0.1875),(0.5625,0.4375),(0.5625,0.5625),(0.5625,0.6875),(0.5625,0.8125),(0.5625,0.9375),(0.625,0.25),(0.625,0.375),(0.625,0.625),(0.625,0.75),(0.625,0.875),(0.6875,0.1875),(0.6875,0.4375),(0.6875,0.5625),(0.6875,0.6875),(0.6875,0.8125),(0.75,0),(0.75,0.375),(0.75,0.5),(0.75,0.625),(0.75,0.75),(0.8125,0.1875),(0.8125,0.4375),(0.8125,0.5625),(0.875,0.25),(0.875,0.375),(0.9375,0.1875),(1,0)}}{%
\begin{scope}[xshift=\xs cm,x=2.72cm,y=0.88cm]
 \fill[black!8] (0,0)--(0.25,0.75)--(0.5,1)--(0.75,0.75)--(1,0)--cycle;
 \draw[black!30,thin] (0,0)--(0.25,0.75)--(0.5,1)--(0.75,0.75)--(1,0)--cycle;
 \draw[->,black!60,thin] (0,0)--(1.06,0) node[right,font=\scriptsize]{$p_1$};
 \draw[->,black!60,thin] (0,0)--(0,1.12) node[above,font=\scriptsize]{$W_{12}$};
 \foreach \pt in \pts {\fill \pt circle (1.1pt);}
 \node[below=1mm,font=\scriptsize] at (0.5,-0.08) {$m=\m$};
\end{scope}}
\end{tikzpicture}
\caption{The $(p_1,W_{12})$-projection of $\cS_2(mK_4)$ (dots, exact)
inside $\operatorname{conv}\cS_2(K_4)$ (shaded): the $m$-fold Minkowski
average fills the hull at rate $\Theta(K/n)=\Theta(1/m)$
(Theorem~\ref{thm:hyperfinite-convexification}
and Remark~\ref{thm:sharp-convexification}).}
\label{fig:k4-convexification}
\end{figure}

\paragraph{Proof sketch and a concrete rate.} The proof writes the quotient
set of the decomposed graph as a vertex-weighted Minkowski average of its
component quotient sets and applies a weighted Shapley--Folkman lemma in
dimension $r_q$; comparing the original and deleted graphs contributes
$4\varepsilon$ across the two Hausdorff legs
(Appendix~\ref{app:proof-hyperfinite-convexification}).  For $P_n$ or $C_n$, deleting
$n/K+1$ edges leaves paths of order $K$, so the bound is
$4(\frac1K+\frac1n)+\frac{6r_qK}n$; taking $K=\lceil\sqrt n\rceil$
gives an explicit $O_q(n^{-1/2})$ rate.
Figure~\ref{fig:k4-convexification} shows the mechanism on $mK_4$.

\paragraph{Why the state-space dimension is necessary.}
For the matching $M_{2m}$, an assignment is exactly a denominator-$m$
probability type on the $N_q=\binom{q+1}{2}$ unordered color pairs.  The
linear map from this type to $(p,W)$ distorts $\ell^1$ distance by a factor
between one and two.  The exact simplex-grid covering radius is
$R_{q,m}$ from \eqref{eq:type-grid-radius}, proving
\eqref{eq:matching-l1-sharpness}.  This is a finite type-lattice calculation,
not an asymptotic appeal to Shapley--Folkman; related quantization problems
are studied by \citet{reznik2011quantization}.

The Euclidean proof behaves differently.  Independently rounding the
component selected in each convex combination makes cross terms vanish, so
the squared error is controlled by $\sum_j(|C_j|/n)^2\leq K/n$.  Applying
this only to the at most $r_q$ exceptional Shapley--Folkman components gives
the other branch of \eqref{eq:hyperfinite-convexification-l2}.  The matching
type lattice attains both regimes.  Thus the $\ell^1$ and $\ell^2$ metrics
give the same fixed-$q$ convergence but genuinely different uniform
finite-$q$ bounds.

\paragraph{Task-specific dimension.}
The dimension factor $r_q=\Theta(q^2)$ in
Theorem~\ref{thm:hyperfinite-convexification} is the price of controlling
\emph{every} coordinate of the quotient at once.  A task that reads only
$k$ coordinates pays for $k$.

\begin{proposition}[Projected convexification]
\label{prop:projected-convexification}
Let $\Pi:\mathbb R^q\times\mathbb R^{q\times q}\to\mathbb R^k$ be linear and
nonexpansive, $\|\Pi z\|_1\leq\|z\|_1$ (for instance the restriction to
any $k$ coordinates of $(p,W)$).  Under the hypothesis of
Theorem~\ref{thm:hyperfinite-convexification},
\begin{equation}
 d_{\rm H}\bigl(\Pi\cS_q(G),\operatorname{conv}\Pi\cS_q(G)\bigr)
 \;\leq\;4\varepsilon+\frac{2(1+D)\,k\,K}{n},
 \label{eq:projected-convexification}
\end{equation}
with $d_{\rm H}$ taken in the $\ell^1$ norm of $\mathbb R^k$.
Consequently, for every $L$-Lipschitz $F:\mathbb R^k\to\mathbb R$,
$\bigl|\min_{z\in\cS_q(G)}F(\Pi z)-\min_{z\in\operatorname{conv}\cS_q(G)}F(\Pi z)\bigr|
 \leq L\bigl(4\varepsilon+2(1+D)kK/n\bigr)$, and the same holds for maxima.
The factor $k$ is sharp before saturation: for $M_{2m}$ and every
$k\leq\min\{N_q-1,2m\}$, some restriction $\Pi$ to $k$ upper-triangular
$W$-coordinates satisfies the following bound; here $n=2m$ and the
component scale is $K=2$:
\begin{equation}
 d_{\rm H}\bigl(\Pi\cS_q(M_{2m}),
  \operatorname{conv}\Pi\cS_q(M_{2m})\bigr)
 \geq\frac{k}{4m}=\frac{k}{4}\frac Kn.
 \label{eq:projected-convexification-lower}
\end{equation}
\end{proposition}

The convexification error $\kappa_q$ in Theorem~\ref{thm:task-value-size-transfer}
and Proposition~\ref{prop:lagrangian-bridge} may therefore be replaced, for
any task that depends on $z$ only through $\Pi z$, by
$\kappa_\Pi(G):=d_{\rm H}(\Pi\cS_q(G),\operatorname{conv}\Pi\cS_q(G))$,
which obeys \eqref{eq:projected-convexification}.  The balanced-cut value
$\Phi$ of \eqref{eq:balanced-cut-query} reads $k=2$ coordinates, the
balance--defect tradeoffs of Appendix~\ref{app:example_intro} read
$k=q+1$, and a homogeneous energy reads $k=1$; for an affine energy no
convexification error enters its optimum, even though $\kappa_\Pi$ itself
need not vanish (Proposition~\ref{prop:convex-sufficiency}).  The proof,
including \eqref{eq:projected-convexification-lower}, is in
Appendix~\ref{app:proof-projected-convexification}.

\paragraph{Locating Theorem~\ref{thm:hyperfinite-convexification}.}
Two qualifiers locate the theorem:
\citet[Example~3.3.7]{borgs2018lp2} observed the disjoint-union case qualitatively, and the theorem supplies the general finite-graph rate, whose $q$- and $K/n$-dependences are matched separately by Remark~\ref{thm:sharp-convexification}; the optimal coefficient of the $\varepsilon$ term remains open;
\citet[Theorem~10.5]{hatami2014localglobal} give a qualitative counterpart
whose input is a local limit rather than a finite decomposition.
One qualifier on Corollary~\ref{cor:hyperfinite-scalar-completeness}: the
classical equivalences \citep{bclsv2012,borgs2018lp2} use
\emph{microcanonical} energies---constrained optima not produced by our
feature map---and are silent at bounded degree
(see ``Normalized quotients and ground-state energies'' in
Appendix~\ref{app:extended-related}).  In their regime they show that the
unconstrained family is insufficient in general; the corollary identifies
hyperfiniteness as a broad regime where the implementable family suffices,
and Corollary~\ref{cor:necessity} shows that the hypothesis cannot be dropped
without replacement.

\paragraph{Growing-component instances.}
Taking $X=K_{D+1}$ (where $h_0\geq1$) gives $\tfrac12K/n$ with $K=D+1$;
$D$-regular expanders with $h_0$ bounded below exist for every fixed
$D\geq3$ and arbitrarily large $K$ \citep{bollobas1988isoperimetric},
giving $c(h_0)K/n$ with $K$ unbounded.  The linear dependence on $K/n$ in
\eqref{eq:hyperfinite-convexification} is therefore sharp even when $q=2$;
the matching clause of Remark~\ref{thm:sharp-convexification} separately
makes the $r_q=\Theta(q^2)$ dependence sharp when $K=2$.

\paragraph{An explicit separating pair.}
On expanders two graphs can be indistinguishable to every
energy-based feature yet differ by a constant on a natural task; no amount
of scalar-energy information extends $\Phi$ continuously, and
Appendix~\ref{app:why-convexity} works out an explicit pair, $K_4$
versus $K_4\sqcup K_4$, with its geometry.

\paragraph{The Lagrangian bridge as a duality-gap estimate.}
Each Lagrangian approximation in
Proposition~\ref{prop:lagrangian-bridge} is an unconstrained energy shifted by a
constant, so constrained optima are recoverable from implementable
measurements up to the convexification error and no further: the classical
Shapley--Folkman duality-gap estimate
\citep{aubin1976estimates,udell2016bounding} instantiated on quotient sets
and calibrated on both sides by
Theorem~\ref{thm:hyperfinite-convexification} and
Proposition~\ref{prop:expander-separation}.
Appendix~\ref{app:why-convexity} gives the
in-depth discussion and Appendices~\ref{app:proof-hyperfinite-convexification},
\ref{app:proof-expander-separation}, and~\ref{app:proof-lagrangian-bridge}
the proofs.

\paragraph{The Potts family.}
Take $a=0$ and $B=I_q$.  Since
$\tfrac12\langle I_q,W(G,c)\rangle
 =\tfrac1n|\{uv\in E(G):c(u)=c(v)\}|$,
the energy \eqref{eq:pairwise-factorization} is the density of
monochromatic edges---the Hamiltonian of the $q$-state Potts
antiferromagnet.  Its ground-state energy is $0$ exactly when $G$ admits a
proper $q$-coloring, so the model detects $\chi(G)\le q$, and the
zero-temperature partition function is the chromatic polynomial at $q$
\citep{fortuin1972,sokal2005}.
The example is the Hamiltonian
$H(c)=-J\sum_{uv\in E(G)}\delta_{c(u)c(v)}$ of the $q$-state Potts model,
where $\delta_{ij}$ is the Kronecker delta, at
coupling $J=-1$ and zero external field \citep{potts1952,wu1982}; a nonzero
field $h\in\mathbb R^q$ is recovered as the unary cost $a=-h$.  Since the
bichromatic density is $\bar d/2$ minus the monochromatic density, the same
ground state solves \textsc{Max-$q$-Cut} \citep{frieze1997}; reversing the
sign to $B=-I_q$ gives the ferromagnet, whose ground states are the
assignments that are constant on each connected component.

\paragraph{Necessity of the hyperfinite hypothesis.}
The hyperfiniteness hypothesis of
Corollary~\ref{cor:hyperfinite-scalar-completeness} cannot simply be
dropped.

\begin{corollary}[Scalar energies are not complete in general]
\label{cor:necessity}
There is a growing $D$-bounded sequence $(H_i)$ such that every homogeneous
finite-state unary--pairwise ground-state energy converges, yet $(H_i)$ is
not partition convergent.
\end{corollary}

The proof (Appendix~\ref{app:proof-necessity}) alternates a
partition-convergent expander sequence with its own two-copy disjoint
unions.  Whether the quantifier over all $q$ in
Corollary~\ref{cor:hyperfinite-scalar-completeness} can be bounded---for
example, whether the two-state energies alone suffice---remains open.

\paragraph{The Lagrangian bridge in full.}
The formal version of Proposition~\ref{prop:lagrangian-bridge} reads as
follows.

\begin{proposition}[Lagrangian bridge; formal]
\label{prop:lagrangian-bridge-formal}
Let $G$ be a finite $D$-bounded graph, $B$ a symmetric interaction matrix,
$a\in\Delta_q$ a target balance, and $L\geq0$ a penalty level.  Write
\begin{equation}
 \begin{aligned}
 \mathsf m_{B,L,a}(G)
 &:=\min_{(p,W)\in\cS_q(G)}
   \Bigl(\tfrac12\langle B,W\rangle+L\|p-a\|_1\Bigr),\\
 \mathsf g_{B,a}(u)
 &:=\min_{(p,W)\in\cS_q(G)}
   \Bigl(\tfrac12\langle B,W\rangle+\langle u,p-a\rangle\Bigr).
 \end{aligned}
 \label{eq:penalized-microcanonical}
\end{equation}
for the penalized microcanonical value and its Lagrangian dual at
multiplier $u\in\mathbb R^q$; each $\mathsf g_{B,a}(u)$ is an unconstrained
unary--pairwise ground-state energy shifted by the constant
$\langle u,a\rangle$.  Then
\begin{equation}
 0\;\leq\;
 \mathsf m_{B,L,a}(G)-\max_{\|u\|_\infty\leq L}\mathsf g_{B,a}(u)
 \;\leq\;
 \max\{L,\tfrac12\|B\|_{\max}\}\;\kappa_q(G).
 \label{eq:duality-gap}
\end{equation}
\end{proposition}  The content of the
proposition beyond the classical duality-gap estimates
\citep{aubin1976estimates,udell2016bounding} is the instantiation to
quotient-set geometry: the bound is in the quotient metric, uniform over
$(B,a)$ at fixed Lipschitz level, and calibrated on both sides.  On any
$(\varepsilon,K)$-decomposable graph,
Theorem~\ref{thm:hyperfinite-convexification} makes the right-hand side
explicit, so unconstrained energies recover every penalized microcanonical
value up to that error; this is the precise sense in which
Corollary~\ref{cor:hyperfinite-scalar-completeness} strengthens
microcanonical-energy characterizations on hyperfinite classes
(Section~\ref{sec:related-work}).  Conversely, on an expander the gap is
real: with $q=2$, $B$ the off-diagonal indicator, $a=(\frac12,\frac12)$, and
$L=\frac12$, the primal value is $\Phi(G)\geq\min\{1/2,h/2\}$ by
Proposition~\ref{prop:expander-separation} while the dual is $0$, so
\eqref{eq:duality-gap} certifies $\kappa_2(G)\geq\min\{1,h\}$ for every
graph with edge expansion $h$---an $\Omega(1)$ convexification error
complementing the $O(K/n)$ regime.

\subsection{Deferred material for Section~\ref{sec:examples}}
\label{app:deferred-examples}

\paragraph{Relaxed edge orthogonality.}
The finite-state calculus is no restriction: continuous spin
spaces enter through finite nets at a cost controlled by the degree bound
alone (Proposition~\ref{prop:compact-spin}).  A concrete instance is the unit
sphere with repulsive interaction $h(x,y)=\langle x,y\rangle^2$, capturing
the orthogonal representations behind the Lov\'asz $\vartheta$ function
\citep{lovasz1979shannon}.  The relaxed edge-orthogonality energy is
\begin{equation}
 \varepsilon_d^\perp(G)
 :=\frac1n\min_{x_v\in S^{d-1}}
 \sum_{uv\in E(G)}\langle x_u,x_v\rangle^2.
 \label{eq:relaxed-edge-orthogonality}
\end{equation}
On a finite graph it vanishes exactly when adjacent vertices can be assigned
orthogonal real unit vectors in dimension $d$.  Thus ordinary coloring and
edge-orthogonal vector assignments are two zero-temperature repulsive spin
models in the same quotient calculus. Learned edge-orthogonal embeddings of this kind have been used for approximate coloring, together with Lovász-$\vartheta$ colorability certificates~\citep{le2026contrastive}.

\subsection{Deferred material for Section~\ref{sec:size-transfer}}
\label{app:deferred-size-transfer}

\paragraph{A finite probe net.}
Theorem~\ref{thm:task-value-size-transfer} turns a family of scalar labels
into a certificate for every continuous quotient task on decomposable
graphs.  It also has a finite version.  If $\mathcal N_\rho$ is a finite
$\rho$-net, in the norm $\|\cdot\|_*$, of the unit ball
$\{\theta:\|\theta\|_*\leq1\}$, then
\begin{equation}
 \eta_q(G,G')
 \leq
 \max_{\theta\in\mathcal N_\rho}
 |\gamma_\theta(G)-\gamma_\theta(G')|
 +2(1+D)\rho.
 \label{eq:finite-probe-transfer}
\end{equation}
Hence, at any fixed accuracy, a finite vector of normalized pairwise
ground-state energies approximately controls every Lipschitz $q$-state
quotient query with a prescribed Lipschitz constant, and each continuous
query through its own modulus of continuity.  The vector is independent of
graph order, although computing its coordinates can be hard.
Support-function duality identifies $\eta_q$ with the Hausdorff distance
between the two convexified quotient sets.

\paragraph{Training datasets and probe covers.}
In general the training dataset of
Corollary~\ref{cor:finite-bank-size-generalization} is existential:
compactness supplies neither an efficient construction nor a useful bound on
$n_{\max}$.  On classes with $\kappa_{q_j}\leq\bar\kappa$, equation
\eqref{eq:finite-probe-transfer} turns a $b$-cover by finite probe vectors
into a quotient cover of radius $b+2(1+D)\rho+2\bar\kappa$; thus
Theorem~\ref{thm:hyperfinite-convexification} connects the certificate to
the small-component regime.

\paragraph{Where the representational ceiling applies.}
The decomposability hypothesis does real work for
\emph{constrained} tasks: relaxation-based learned combinatorial optimization
is often evaluated on random-regular benchmarks
\citep{schuetz2022combinatorial}.  On expander instances,
Proposition~\ref{prop:lagrangian-bridge} certifies an $\Omega(1)$ gap
between constrained and unconstrained optima on graph \emph{pairs} such
as $(G,G\sqcup G)$.  Two qualifications keep this from being a statement
about the benchmarks themselves.  First, the tasks those pipelines
optimize (MAX-CUT, independence) are affine, so energy-based features
represent them exactly on every graph (Proposition~\ref{prop:convex-sufficiency})
and their transfer is governed by $\eta_q$ alone
(Corollary~\ref{cor:opening-task-transfer}); the obstruction there is
optimization, not representation.  Second, the ceiling is a worst case
over pairs with equal hulls and different constrained values; a
\emph{distribution} over connected random $D$-regular graphs need not
contain such a pair.  Almost-sure LD convergence of random regular graphs
is posed as a conjecture by \citet{borgs2017ld}, not used here.  The ceiling bites when training and
test laws place mass on graphs with equal hulls and different constrained
values---e.g.\ connected and disconnected instances mixed, or families
with different expansion at equal energies---and the guarantees of
Section~\ref{sec:size-transfer} for constrained tasks target planar,
lattice-like, and minor-free families instead.

\paragraph{Generalization for linear probe readouts.}
Theorem~\ref{thm:generalization} covers every Lipschitz predictor; the
practical class used in Appendix~\ref{app:experiment-details} admits a
dimension-mild bound (proof in Appendix~\ref{app:proof-generalization}, with
the proof of Theorem~\ref{thm:generalization}).

\begin{proposition}[Generalization for linear probe readouts]
\label{prop:linear-generalization}
Let $B,W_1>0$, $\delta\in(0,1)$, and $m\in\mathbb N$.  Let $\mu$ be a
distribution on finite $D$-bounded graphs with measurable labels
$y(G)\in[0,B]$, draw $G_1,\ldots,G_m$ independently from $\mu$, and write
$\widehat{\mathbb E}_m g:=m^{-1}\sum_{i=1}^m g(G_i)$.  Let $\mathcal N$ be a
set of $P$ probes satisfying $\|\theta\|_*\leq1$.  For the class of
$\ell^1$-bounded linear readouts
$\mathcal F_{\rm lin}
 :=\{G\mapsto\langle w,(\gamma_\theta(G))_{\theta\in\mathcal N}\rangle+b:
 \|w\|_1\leq W_1,\ |b|\leq B\}$, with probability at least $1-\delta$,
\begin{equation}
 \sup_{f\in\mathcal F_{\rm lin}}
 \bigl|\,\mathbb E\,|f-y|-\widehat{\mathbb E}_m|f-y|\,\bigr|
 \;\leq\;
 2W_1(1{+}D)\sqrt{\tfrac{2\ln(2P)}{m}}
 +\tfrac{2B}{\sqrt m}
 +3\bigl(W_1(1{+}D)+2B\bigr)\sqrt{\tfrac{\ln(4/\delta)}{2m}}.
 \label{eq:linear-generalization}
\end{equation}
\end{proposition}

\paragraph{Generalization over the quotient state.}
Section~\ref{sec:size-transfer} summarizes the following covering-number
bound in words; the proof is in Appendix~\ref{app:proof-generalization}.

For fixed state sizes $q_1,\ldots,q_s$, write
$\mathfrak X_s:=\prod_{j=1}^s\cK(\mathsf Y_{D,q_j})$, equipped with the
max metric $d_R$ of Section~\ref{sec:size-transfer}; thus
$R(G)=(\cS_{q_1}(G),\ldots,\cS_{q_s}(G))\in\mathfrak X_s$.

\begin{theorem}[Generalization over the quotient state]
\label{thm:generalization}
Let $B,\Lambda>0$, $\delta\in(0,1)$, and $m\in\mathbb N$.  Let $\mu$ be any
distribution on finite $D$-bounded graphs with measurable labels
$y(G)\in[0,B]$, and draw $G_1,\ldots,G_m$ independently from $\mu$.  With probability at least
$1-\delta$, uniformly over all $\Lambda$-Lipschitz
$f:\mathfrak X_s\to[0,B]$,
\begin{equation}
\begin{aligned}
 &\bigl|\,\mathbb E\,|f(R(G))-y(G)|
 -\tfrac1m\textstyle\sum_i|f(R(G_i))-y(G_i)|\,\bigr|\\
 &\qquad\leq
 \inf_{0<\rho\leq1+D}\Bigl[6\Lambda\rho
 +B\sqrt{\tfrac{\mathsf N_{\mathfrak X}(\rho)\ln(3+\frac{B}{\Lambda\rho})
 +\ln\frac2\delta}{2m}}\Bigr].
\end{aligned}
 \label{eq:generalization-bound}
\end{equation}
where $\mathsf N_{\mathfrak X}(\rho)
\leq\prod_{j\leq s}2^{\,(3(1+D)/\rho)^{q_j+q_j^2}}$ is the
$\rho$-covering number of $\mathfrak X_s$.
\end{theorem}

\paragraph{Generalization over the probe representation.}
The hyperspace covering number in Theorem~\ref{thm:generalization} is
doubly exponential in $q^2$.  The finite probe representation of
\eqref{eq:finite-probe-transfer} lives in a cube of dimension $P=|\mathcal N|$,
whose covering number is only polynomial in $1/\rho$ with exponent $P$,
and by \eqref{eq:probe-quotient-sandwich} the probe map
$\gamma:G\mapsto(\gamma_\theta(G))_{\theta\in\mathcal N}$ is
$1$-Lipschitz from $(\cK(\mathsf Y_{D,q}),d_{\rm H})$ to
$(\mathbb R^P,\|\cdot\|_\infty)$, so every Lipschitz predictor of the
probe vector is a Lipschitz predictor of the quotient state.

\begin{proposition}[Generalization over the probe representation]
\label{prop:probe-generalization}
Let $\mathcal N$ be a finite set of $P$ probes with $\|\theta\|_*\leq1$, so
that $\gamma(G)\in[-(1{+}D),1{+}D]^P$.  In the setting of
Theorem~\ref{thm:generalization}, with probability at least $1-\delta$,
uniformly over all $f:[-(1{+}D),1{+}D]^P\to[0,B]$ that are
$\Lambda$-Lipschitz for $\|\cdot\|_\infty$,
\begin{equation}
 \begin{aligned}
 &\bigl|\,\mathbb E\,|f(\gamma(G))-y(G)|
 -\tfrac1m\textstyle\sum_i|f(\gamma(G_i))-y(G_i)|\,\bigr|\\
 &\qquad\leq
 \inf_{0<\rho\leq1+D}\Bigl[6\Lambda\rho
 +B\sqrt{\tfrac{\mathsf N_P(\rho)\ln(3+\frac{B}{\Lambda\rho})
 +\ln\frac2\delta}{2m}}\Bigr].
 \end{aligned}
 \label{eq:probe-generalization}
\end{equation}
where $\mathsf N_P(\rho)\leq(2(1+D)/\rho)^{P}$ is the $\ell^\infty$
$\rho$-covering number of the probe cube.
\end{proposition}

The bound is the analogue of \eqref{eq:generalization-bound} with the
hyperspace replaced by the probe cube: the exponent $q+q^2$ inside a
double exponential becomes the single exponent $P$.  It is still a
covering bound and is slower than Proposition~\ref{prop:linear-generalization}
for linear readouts, but it is the operative uniform bound for
\emph{nonlinear} readouts of the probe vector, e.g.\ the heads following
Theorem~\ref{thm:raw-rni-encoder}.  The proof (Appendix~\ref{app:proof-generalization})
is that of Theorem~\ref{thm:generalization} with the state space replaced
by the cube.

\paragraph{On the generalization bounds.}
In Theorem~\ref{thm:generalization}, the hyperspace covering number is
enormous, exactly as in the graphon-signal analysis it mirrors; the honest
summary is that \eqref{eq:generalization-bound} is explicit and
distribution-free but astronomically slow, while
\eqref{eq:linear-generalization}---the class used in
Appendix~\ref{app:experiment-details}---is dimension-mild.  On classes with
$\kappa_q\leq\bar\kappa$ (Theorem~\ref{thm:hyperfinite-convexification}),
two graphs with the same finite $\rho$-net probe vector have quotient
distance at most $2(1+D)\rho+2\bar\kappa$.  Thus any target that is
$\Lambda$-Lipschitz in quotient Hausdorff distance differs across such a fiber by at most
$2\Lambda((1+D)\rho+\bar\kappa)$.  This is the representation-bias term on
the stated small-convexification regimes; it is separate from the
$\kappa_q$-free statistical complexity bounds above.

\subsection{When and why convexification is necessary}
\label{app:why-convexity}

For readers who think in transferability and generalization bounds, the
convexification error $\kappa_q$ has a fixed address.  Decompose the test
error of a size-transferring predictor as
\[
 \text{test error}
 \;\leq\;
 \underbrace{\text{training error}}_{\text{optimization}}
 \;+\;
 \underbrace{\text{generalization gap}}_{\text{Theorem~\ref{thm:generalization}: }\kappa_q\text{-free}}
 \;+\;
 \underbrace{\text{representation bias}}_{\text{convexification }\kappa_q
   \,+\,\text{probe/encoder approximation}}.
\]
The reason $\kappa_q$ is the unavoidable part of the third term is a ceiling
on energy features.  Each
feature coordinate is a probe energy, hence a minimum of an affine functional
over $\cS_q(G)$; any downstream readout factors through this probe vector and
therefore through the convex hull (Proposition~\ref{prop:convex-sufficiency}).
Theorem~\ref{thm:raw-rni-encoder} approximates these coordinates on raw
bounded-treewidth graphs.  The hull is what energy features \emph{are}; the
nonconvex set is what tasks \emph{need};
$\kappa_q=d_{\rm H}(\cS_q,\operatorname{conv}\cS_q)$ is the gap between the
two.  Accordingly $\kappa_q$ enters the separate target-versus-representation
comparison whenever the pipeline replaces the state by energy measurements---the right half
of \eqref{eq:probe-quotient-sandwich}, the rounding radius
\eqref{eq:rounding}, the certificate slack in
\eqref{eq:effective-certificate}---but not the statistical complexity bound
\eqref{eq:linear-generalization}, which is explicitly $\kappa_q$-free.
{It also does not enter affine-task bounds stated directly in terms of
the energy/hull distance $\eta_q$, such as
Corollary~\ref{cor:opening-task-transfer}.}

\begin{figure}[t]
\centering
\begin{tikzpicture}[>=Latex,every node/.style={font=\scriptsize}]
\foreach \panel/\xs/\pts in {%
 a/0/{(0,0),(0.25,0.75),(0.5,1),(0.75,0.75),(1,0)},%
 b/5.2/{(0,0),(0.125,0.375),(0.25,0.5),(0.25,0.75),(0.375,0.375),(0.375,0.875),(0.5,0.75),(0.5,1),(0.625,0.375),(0.625,0.875),(0.75,0.5),(0.75,0.75),(0.875,0.375),(1,0)}}{%
\begin{scope}[xshift=\xs cm,x=3.4cm,y=1.5cm]
 \fill[black!8] (0,0)--(0.25,0.75)--(0.5,1)--(0.75,0.75)--(1,0)--cycle;
 \draw[black!30,thin] (0,0)--(0.25,0.75)--(0.5,1)--(0.75,0.75)--(1,0)--cycle;
 \draw[->,black!60,thin] (0,0)--(1.08,0) node[right] {$p_1$};
 \draw[->,black!60,thin] (0,0)--(0,1.12) node[above] {$W_{12}$};
 \foreach \pt in \pts {\fill \pt circle (1.3pt);}
\end{scope}}
\begin{scope}[x=3.4cm,y=1.5cm]
 \node[star,star points=5,star point ratio=2.4,draw,inner sep=1.1pt]
   at (0.5,0) {};
 \node at (0.5,-0.28) {(a) $\cS_2(K_4)$: $\star=(\tfrac12,0)$ in hull only};
 \begin{scope}[shift={(0.895,0.93)},x=1cm,y=1cm]
  \foreach \i/\gx/\gy in {1/-0.11/0.11,2/0.11/0.11,3/0.11/-0.11,4/-0.11/-0.11}
    {\coordinate (g\i) at (\gx,\gy);}
  \draw[black!60,very thin] (g1)--(g2)--(g3)--(g4)--(g1) (g1)--(g3) (g2)--(g4);
  \foreach \i in {1,...,4}{\fill[black!70] (g\i) circle (0.9pt);}
 \end{scope}
\end{scope}
\begin{scope}[xshift=5.2cm,x=3.4cm,y=1.5cm]
 \fill (0.5,0) circle (1.3pt);
 \draw (0.5,0) circle (3.2pt);
 \node at (0.5,-0.28) {(b) $\cS_2(K_4\sqcup K_4)$: $(\tfrac12,0)$ attained (circled)};
 \begin{scope}[shift={(0.87,0.93)},x=1cm,y=1cm]
  \foreach \s in {-0.17,0.17}{
   \foreach \i/\hx/\hy in {1/-0.08/0.08,2/0.08/0.08,3/0.08/-0.08,4/-0.08/-0.08}
     {\coordinate (h\i) at ({\s+\hx},\hy);}
   \draw[black!60,very thin] (h1)--(h2)--(h3)--(h4)--(h1) (h1)--(h3) (h2)--(h4);
   \foreach \i in {1,...,4}{\fill[black!70] (h\i) circle (0.8pt);}
  }
 \end{scope}
\end{scope}
\end{tikzpicture}
\caption{The feature blind spot, in the
$(p_1,W_{12})$-projection with exact coordinates.  (a)~$\cS_2(K_4)$ (dots)
and its convex hull (shaded); the balanced zero-cut target
$(p_1,W_{12})=(\frac12,0)$ ($\star$) lies in the hull---as the midpoint of
the two monochromatic quotients---but not in the set, and
$\Phi(K_4)=\frac12$.  (b)~$\cS_2(K_4\sqcup K_4)$ (dots): the hull is
\emph{identical} to (a), so every probe energy agrees on the two graphs,
yet the set now contains $(\frac12,0)$ (circled; color one copy state~$1$,
the other state~$2$), and $\Phi(K_4\sqcup K_4)=0$.  Energy features see
only the shaded region and cannot distinguish (a) from (b); the balanced-cut
task lives in the difference between dots and shade.}
\label{fig:k4-blindspot}
\end{figure}
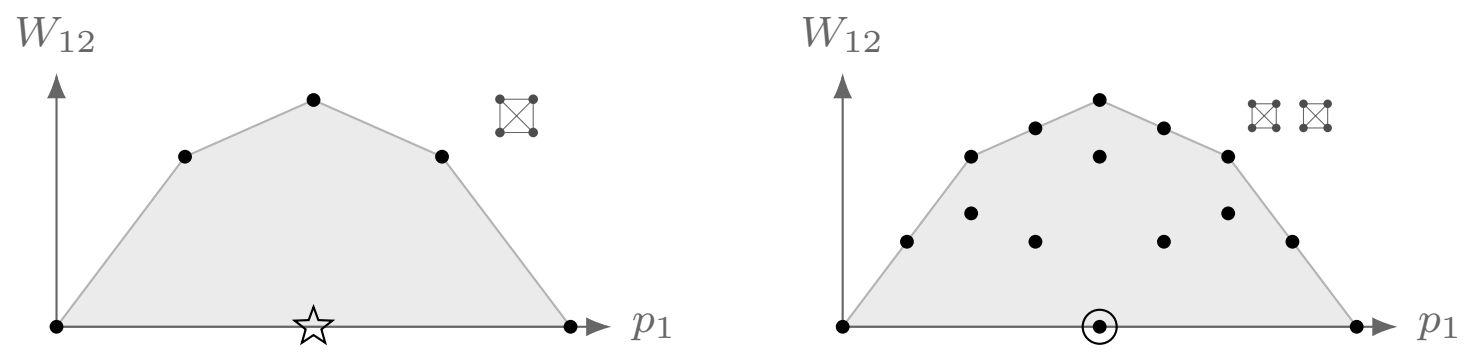

\textbf{A two-graph example.}  Let $G=K_4$ and $G'=K_4\sqcup K_4$.  Vertex
normalization makes every unary--pairwise ground-state energy of the two
graphs \emph{identical}: $\gamma_\theta(G')=\gamma_\theta(G)$ for every
probe $\theta$, by \eqref{eq:expander-convex-equality}.  So every energy
feature map, of any dimension---including any network whose outputs
factor through such energies---assigns both graphs the same feature vector,
and every readout on it the same prediction.
Figure~\ref{fig:k4-blindspot} shows the two quotient sets and their common
convex hull.

Yet the balanced-cut task \eqref{eq:balanced-cut-query}, the objective of
balanced graph partitioning, separates them: $\Phi(G')=0$ (color one copy
state~$1$, the other state~$2$), while $\Phi(K_4)=\tfrac12$ (check
$k=0,\ldots,4$ state-$1$ vertices by hand:
$\min\{\tfrac12,\,\tfrac14+\tfrac34,\,0+1\}=\tfrac12$).  Any predictor of
$\Phi$ over energy features therefore errs by at least $\tfrac14$ on one of
the two graphs---an architecture- and training-independent impossibility of
the transferability kind.  Proposition~\ref{prop:expander-separation} shows
the same gap $\min\{1/2,h/2\}$ persists along growing expander sequences,
so this is not a small-graph artifact, and
Proposition~\ref{prop:lagrangian-bridge} certifies it as
$\kappa_2\geq\min\{1,h\}$.

\textbf{When it is not a problem.}
Theorem~\ref{thm:hyperfinite-convexification} and
Remark~\ref{thm:sharp-convexification} give the universal upper bound
$4\varepsilon+O_D(q^2K/n)$, with the $q^2$ and $K/n$ dependences separately
matched in the worst case.  Thus, after choosing a decomposition with small
$\varepsilon$ and then sending $n\to\infty$, a finite order-independent
energy representation chosen for any prescribed accuracy becomes
asymptotically sufficient for the task state.  A particular graph may have
zero gap, and the decomposition alone
does not certify an impossibility.  The resulting regime map is a three-level feature hierarchy for
size transfer: degree and size statistics fail already on hyperfinite
instances (for example, $C_{120}$ and $40K_3$ both have $120$ vertices
and are $2$-regular, but their coloring defects differ by $1/3$);
probe energies are complete there but
hit the $K_4$-versus-$K_4\sqcup K_4$ ceiling on expanders; and past the
ceiling no scalar-energy feature map suffices---one must carry an approximation of
the nonconvex set itself
(Proposition~\ref{prop:treewidth-encoder})
or solve the constrained problem per instance.  Convexification is thus not
an assumption of the pipeline but a sufficient regime certificate: a good
decomposition proves when energy features suffice asymptotically, while a
nonvanishing measured gap proves when they cannot.

\subsection{Experiment details}
\label{app:experiment-details}

This subsection and Appendix~\ref{app:trained-experiments} give the
protocols for the three representations and models compared in
Section~\ref{sec:size-transfer}.

\paragraph{Graph families and data split.}
Families: cycles $C_n$, paths $P_n$, $2\times k$ ladders, random subcubic
trees (uniform attachment to a vertex of degree $<3$, seeds fixed), and
disjoint triangle unions $mK_3$; all have maximum degree $\leq3$ and
treewidth $\leq2$.  Training set: $72$ graphs with $8\leq n\leq30$; test
set: $66$ graphs with $120\leq n\leq360$.  The dataset-generation seed is
$11$; the random-tree seeds are $1000$--$1009$ in training and
$1100$--$1109$ in testing.  All three models predict
$(e_2,\mathrm{MaxCut}/n,\alpha/n)$ on this same split, and evaluation uses
mean absolute error on the test graphs.

\paragraph{Exact energy features.}
We use $45$ two-state probes $\theta=(a,B)$ drawn once from a standard
Gaussian and normalized to $\|\theta\|_*=1$ (seed $7$); the probes that
reproduce $e_2$, $\mathrm{MaxCut}/n$, and $\alpha/n$ exactly ($B=I$;
$B=-\text{offdiag}$; $a=(-1,0)$, $B=\mathrm{diag}(1,0)$) are excluded, so
the readout must reconstruct the tasks from generic energies.  All
$\gamma_\theta$ values are exact: chain and cyclic transfer-matrix dynamic
programs for paths and cycles, a $4$-state column program for ladders,
rooted dynamic programs for trees, and enumeration per triangle for
$mK_3$.  When \texttt{experiments/toy\_transfer.py} is run as a script,
its main guard first calls a self-check that compares the family solvers
and closed-form task values with brute force on small instances.  The
trained-model driver does not invoke that check automatically.  The
readout is one ridge regression per task ($\lambda=10^{-4}$), with features
standardized using the training-set means and standard deviations.

\paragraph{Spectral--motif ridge baseline.}
The structural features are inverse order $1/n$, mean degree $2|E|/n$,
the degree histogram $(\mathrm{deg}=1,2,3)$, triangle density, and the second
to sixth spectral moments of the normalized adjacency matrix.  We fit one
ridge regression per task with $\lambda=10^{-4}$, again standardizing
features using the training-set statistics.  This baseline receives no
energy probes.

\subsection{Neural baseline}
\label{app:trained-experiments}

GPS--RWSE \citep{rampasek2022graphgps} combines, in each of four width-$64$
blocks, a local GIN branch with four-head global self-attention.  Write $A$
for the adjacency matrix and $D_{\rm deg}$ for the diagonal degree matrix.
Its deterministic node input contains the degree one-hot encoding, $1/n$,
and the return probabilities $((D_{\rm deg}^{-1}A)^t_{vv})_{t=1}^{20}$.
The model uses layerwise mean readout and an MLP head, with $189{,}575$
parameters.  It is trained on the split of
Appendix~\ref{app:experiment-details} by mean squared error on standardized
targets, using full-batch AdamW with learning rate $10^{-3}$, weight decay
$10^{-5}$, and cosine decay for $1500$ epochs.  Reported test MAEs are
means over model seeds $\{0,1,2\}$.

RWSE is global preprocessing, so this comparator is not a raw local MPNN.
Indeed,
\[
 n^{-1}\sum_v(D_{\rm deg}^{-1}A)^t_{vv}
 =n^{-1}\operatorname{Tr}((D_{\rm deg}^{-1/2}AD_{\rm deg}^{-1/2})^t),
\]
so the neural baseline receives the spectral information of the
handcrafted comparator without receiving any energy probe.  The exact
energy features above are computed by family-specific dynamic programs;
this comparison tests the energy representation rather than implementing
the RNI-MPNN encoder of Theorem~\ref{thm:raw-rni-encoder}.

\section{Repair cost, exactness, and richer observations}
\label{sec:boundary}

All objectives in Section~\ref{sec:examples} average a bounded cost over
vertices or edges.  We now quantify the resulting edit stability and show
that exact colorability is excluded by the same mechanism.

\subsection{Disjoint unions and negligible edits}
\label{app:disjoint-unions-edits}

For compact sets $A,B$ in a vector space and $\lambda\in[0,1]$, write
$\lambda A+(1-\lambda)B$ for their weighted Minkowski sum.

\begin{proposition}[Unions and edit stability]
\label{prop:disjoint-union}
For finite graphs $G,H$ and
$\lambda=|V(G)|/(|V(G)|+|V(H)|)$,
\begin{equation}
 \cS_q(G\sqcup H)=\lambda\cS_q(G)+(1-\lambda)\cS_q(H).
 \label{eq:disjoint-union}
\end{equation}
Consequently, if $|V(H_n)|/|V(G_n)|\to0$ and both sequences are
$D$-bounded, then
\begin{equation}
 d_D^{\mathrm{part}}\!\left(
 \Theta_D^{\mathrm{part}}(G_n\sqcup H_n),
 \Theta_D^{\mathrm{part}}(G_n)
 \right)\longrightarrow0.
 \label{eq:sublinear-vertex-edit}
\end{equation}
If $G_n$ and $G_n'$ have the same vertex set and
$|E(G_n)\triangle E(G_n')|=o(|V(G_n)|)$, and both sequences are $D$-bounded,
their partition distance also tends to zero.
\end{proposition}

The proof is given in Appendix~\ref{app:proof-disjoint-union}.

This stability is useful for normalized optimization, but it makes bounded
obstructions invisible.  An odd cycle is the smallest chromatic example:
$\chi(C_{2m+1})=3$, whereas
$e_2(C_{2m+1})=1/(2m+1)\to0$.  The limit is two-colorable after deleting
$o(n)$ edges although no graph in the sequence is exactly two-colorable.

\subsection{Approximate chromatic number}

The odd-cycle example suggests taking the first palette size whose normalized
repair cost vanishes.  To specify the repair model, we call
\begin{equation}
 \chi_{\rm app}(x):=\min\{q\leq D+1:e_q(x)=0\}
 \label{eq:approximate-chromatic-number}
\end{equation}
the \emph{edge-deletion approximate chromatic number} of a partition-limit
point $x$.  Greedy coloring gives $e_{D+1}=0$.  On a growing sequence,
$e_q\to0$ is also equivalent to deleting $o(n)$ vertices before $q$-coloring:
one can cover the monochromatic edges by at most one chosen endpoint per bad
edge, while deleting $s$ vertices removes at most $Ds$ incident edges.

\begin{proposition}[Continuous defects and a lower-semicontinuous threshold]
\label{prop:approximate-chromatic-number}
The function $\chi_{\rm app}$ is lower semicontinuous on
$\mathfrak X_D^{\mathrm{part}}$ and equals $\chi(G)$ on each finite graph
state.  For every defect-resolution scale $A>0$,
\begin{equation}
 \underline\chi_A(x)
 :=1+\sum_{q=1}^{D}\min\{1,Ae_q(x)\}
 \label{eq:continuous-chromatic-lower-bound}
\end{equation}
is continuous, satisfies $\underline\chi_A(G)\leq\chi(G)$ on finite graphs,
and increases pointwise to $\chi_{\rm app}$ as $A\to\infty$.
\end{proposition}

Appendix~\ref{app:proof-approximate-chromatic-number} contains the proof.

Approximate colorability is therefore the continuous-repair information
carried by partition convergence.  Exact colorability cannot be recovered by
a different continuous post-processing of the same state.

\begin{theorem}[Exact-colorability threshold]
\label{thm:exact-boundary}
For every $q\leq D$, the $\{0,1\}$-indicator of exact $q$-colorability has no
continuous extension to $\mathfrak X_D^{\mathrm{part}}$.  Moreover, suppose a
continuous function $U$ satisfies
$U(\Theta_D^{\mathrm{part}}(G))\geq\chi(G)$ for every finite $D$-bounded
graph.  If $x$ is approached by graphs $G_n$ with $|V(G_n)|\to\infty$, then
\begin{equation}
 U(x)\geq D+1.
 \label{eq:no-nontrivial-upper-bound}
\end{equation}
\end{theorem}

Both statements use the same obstruction: append a fixed clique to a growing
graph.  The clique changes exact feasibility but disappears from every
vertex-normalized quotient.

The proof appears in Appendix~\ref{app:proof-exact-boundary}.

Thus continuity preserves a macroscopic repair cost, not a bounded witness.
The same zero-set argument applies to the least dimension $d$ with vanishing
relaxed edge-orthogonality energy: it is lower semicontinuous, while a fixed
edge-orthogonality obstruction can vanish under an $o(n)$-vertex
perturbation.

\paragraph{Beyond pairwise payoffs.}
The quotient keeps only first edge moments: nonlinear one-hop payoffs need
colored-star statistics, finite-range payoffs the corresponding local--global
radius, and observed attributes typed or measure-valued quotients.  None of
these richer observations is part of the pairwise core model.





\section{Omitted proofs}
\label{app:routine-proofs}

Proofs are grouped to follow the sectioning of the main text;
within each group they appear in order of statement.

\subsection{Proofs for Section~\ref{sec:attainable-regions}}

The only deferred claim of the section is the compactness of the hyperspace
packaging.
\subsubsection{Proof of Theorem~\ref{thm:compactness}}
\label{app:proof-compactness}

\begin{proof}
Each host $\mathsf Y_{D,q}$ is a compact finite-dimensional set, so its
Hausdorff hyperspace $\cK(\mathsf Y_{D,q})$ is compact metrizable.  Tychonoff's
theorem makes the countable product compact, and
\eqref{eq:partition-metric} metrizes it.  The finite-realizable closure is a
closed subspace.
\end{proof}

\subsection{Proofs for Section~\ref{sec:task-calculus}}

The section defers the continuity theorem, the support-function
characterization, the convexification theorem with its corollary, the sharp
norm comparison, the separation proposition, the necessity corollary of
Appendix~\ref{app:deferred-task}, and the Lagrangian bridge; the proofs
follow in that order.
\subsubsection{Proof of Theorem~\ref{thm:task-completeness}}
\label{app:proof-task-completeness}

\begin{proof}
Every continuous map on the compact host $\mathsf Y_{D,q}$ is uniformly
continuous.  If $d_{\rm H}(K_n,K)\leq\delta$, each point of either set lies
within $\delta$ of the other, so their images lie within the corresponding
modulus of continuity.  This proves
\eqref{eq:continuous-range-transfer}, and the real-valued claim follows by
taking the endpoints of the compact images.

For the converse, let $C(\mathsf Y_{D,q})$ denote the real-valued continuous
functions on the host and define
\begin{equation}
 \mathcal U(K):=(\min_KF)_{F\in C(\mathsf Y_{D,q})}.
 \label{eq:query-embedding}
\end{equation}
Every coordinate of $\mathcal U$ is continuous by the first step.  The map is
injective: if $x$ belongs to one of two distinct compact sets but not the
other, then $F(y)=d_q(x,y)$ has minimum zero on the first set and a strictly
positive minimum on the second.  A continuous injection from the compact
hyperspace into a Hausdorff product is a topological embedding.  Therefore
the initial topology induced by these minimum queries is exactly the
Hausdorff topology.
\end{proof}

\subsubsection{Proof of Proposition~\ref{prop:convex-sufficiency}}

\begin{proof}
Equation~\eqref{eq:pairwise-factorization} ranges over every affine functional
of $(p,W)$, and such a functional has the same extrema on a set and its convex
hull.  Support functions uniquely determine compact convex sets.  Because the
host is bounded and finite dimensional, these support functions share a
Lipschitz constant on the unit sphere; pointwise convergence is therefore
uniform, which is equivalent to Hausdorff convergence of the convex sets.
\end{proof}

\subsubsection{Proof of Theorem~\ref{thm:hyperfinite-convexification}}
\label{app:proof-hyperfinite-convexification}

\begin{proof}
Let $H$ be the graph after the edge deletions, with components
$C_1,\ldots,C_m$ and weights $\lambda_j:=|V(C_j)|/n$.  The disjoint-union
identity gives
\begin{equation}
 \cS_q(H)=\sum_{j=1}^m\lambda_j\cS_q(C_j).
 \label{eq:component-minkowski-sum}
\end{equation}
Choose one point from each $\cS_q(C_j)$ and translate that set by its chosen
point.  Every translated set lies in the same linear space: the $p$-coordinates
sum to zero, $W$ is symmetric, and the $W$-coordinates sum to zero.  Its
dimension is at most $r_q$.  Apply the weighted Shapley--Folkman lemma
recalled in Appendix~\ref{app:deferred-task} in this space.  Every point of the convex hull of the sum can be represented with all
but at most $r_q$ summands chosen from the original sets.  Each exceptional
summand has weight at most $K/n$, and the diameter of
$\mathsf Y_{D,q}$ under $d_q$ is at most $2(1+D)$.  Therefore
\begin{equation}
 d_{\rm H}\!\left(\cS_q(H),\operatorname{conv}\cS_q(H)\right)
 \leq\frac{2(1+D)r_qK}{n}.
 \label{eq:component-convexification}
\end{equation}
Using the same assignment before and after the deletion shows
$d_{\rm H}(\cS_q(G),\cS_q(H))\leq2\varepsilon$; taking convex hulls is
nonexpansive in Hausdorff distance.  The triangle inequality yields
\eqref{eq:hyperfinite-convexification}.

For a hyperfinite sequence, choose $K=K(\varepsilon)$ first and then send
$n\to\infty$.  The limit superior is at most $4\varepsilon$, and sending
$\varepsilon\downarrow0$ proves the final claim.
\end{proof}

\subsubsection{Proof of Corollary~\ref{cor:hyperfinite-scalar-completeness}}

\begin{proof}
The forward implication follows from Theorem~\ref{thm:task-completeness}.
Conversely, Proposition~\ref{prop:convex-sufficiency} gives Hausdorff
convergence of $\operatorname{conv}\cS_q(G_n)$ for every $q$, while
Theorem~\ref{thm:hyperfinite-convexification} makes its distance from
$\cS_q(G_n)$ tend to zero.
\end{proof}

\subsubsection{Proof of the sharpness claims in Remark~\ref{thm:sharp-convexification}}
\label{app:proof-sharp-convexification}

\begin{proof}
The proof has three parts: an exact type-lattice calculation for matchings in
$\ell^1$, a growing-component lower bound, and Euclidean rounding with its
matching lower example.

\emph{Matching reduction.}
Index the $N=N_q$ unordered color pairs by $a\in[N]$, and let
\[
 \mathcal T_{N,m}:=\{x\in\Delta_N:mx\in\mathbb Z^N\}
\]
be the denominator-$m$ type grid.  If $x_{ij}$ is the fraction of matching
edges whose endpoint colors form $\{i,j\}$, define the symmetric matrix
$B(x)$ by $B(x)_{ii}=x_{ii}$ and
$B(x)_{ij}=B(x)_{ji}=x_{ij}/2$ for $i<j$, and put
$L(x):=(B(x)\mathbf1,B(x))$.  Directly from the normalization in
\eqref{eq:vertex-proportion},
\begin{equation}
 \cS_q(M_{2m})=L(\mathcal T_{N,m}),
 \qquad
 \operatorname{conv}\cS_q(M_{2m})=L(\Delta_N).
 \label{eq:matching-type-map}
\end{equation}
Moreover
$\|B(x)-B(y)\|_1=\|x-y\|_1$ and
$\|(B(x)-B(y))\mathbf1\|_1\leq\|x-y\|_1$, whence
\begin{equation}
 \|x-y\|_1\leq d_q(Lx,Ly)\leq2\|x-y\|_1.
 \label{eq:type-map-l1-distortion}
\end{equation}

It remains to compute the grid radius.  Given $x\in\Delta_N$, write
$mx_i=a_i+f_i$ with $a_i\in\mathbb Z_{\geq0}$ and $f_i\in[0,1)$.  Then
$s:=\sum_i f_i$ is an integer in
$[0,\min\{m,N-1\}]$.  The closest denominator-$m$ type rounds up the $s$
largest fractional parts and rounds down the rest, at $\ell^1$ cost
\begin{equation}
 \frac2m\left(s-\sum_{i=1}^s f_{(i)}\right)
 \leq\frac{2s(N-s)}{Nm},
 \label{eq:type-rounding-cost}
\end{equation}
where $f_{(1)}\geq\cdots\geq f_{(N)}$ and the inequality uses
$\sum_{i=1}^s f_{(i)}\geq s^2/N$.  Equality is attained by taking every
$f_i=s/N$ and choosing nonnegative integers $a_i$ summing to $m-s$.
Maximizing over $s$ gives $R_{q,m}$ in \eqref{eq:type-grid-radius}.
Equations~\eqref{eq:matching-type-map}--\eqref{eq:type-map-l1-distortion}
prove \eqref{eq:matching-l1-sharpness}; its asymptotic form follows from
$N_q=\Theta(q^2)$.

\emph{Growing components.}
Write $A_1$ for the all-state-$1$ quotient of $G=mX$ and $A_2$ for the
all-state-$2$ quotient, and put $z^*:=(1-t)A_2+tA_1$ with $t=1/(2m)$.
For any binary assignment, let $k_j$ be the number of state-$1$ vertices in
the $j$th copy, $\delta_j:=\min\{k_j,K-k_j\}$, and
$\operatorname{cut}_j$ its number of bichromatic edges.  Then
\[
 p_1=\frac1m\sum_j\frac{k_j}{K},
 \qquad
 W_{12}=\frac1m\sum_j\frac{\operatorname{cut}_j}{K}
 \geq\frac{h_0}{mK}\sum_j\delta_j.
\]
Rounding each $k_j/K$ to $0$ or $1$ moves $p_1$ by at most
$(mK)^{-1}\sum_j\delta_j$, while $t$ is $1/(2m)$ from the grid
$m^{-1}\mathbb Z$.  If this rounding cost is at most $1/(4m)$, the two
$p$-coordinates contribute at least $1/(2m)$ to $d_q$; otherwise the two
off-diagonal $W$-coordinates contribute more than $h_0/(2m)$.  Taking the
distance from $z^*$ to $\cS_2(G)$ proves
\eqref{eq:component-lower-bound}.

\emph{Euclidean upper bound.}
Let $H$ be the decomposed graph, with components $C_j$ and
$\lambda_j=|C_j|/n$.  For a point
$z=\sum_j\lambda_jz_j$ in
$\operatorname{conv}\cS_q(H)$, independently choose
$X_j\in\cS_q(C_j)$ with $\mathbb E X_j=z_j$.  Every quotient vector has
Euclidean norm at most $\sqrt{1+D^2}$, so independence gives
\[
 \mathbb E\left\|\sum_j\lambda_j(X_j-z_j)\right\|_2^2
 \leq(1+D^2)\sum_j\lambda_j^2
 \leq(1+D^2)K/n.
\]
Some realization therefore lies within the square root of this bound.
Alternatively, first apply Shapley--Folkman and let $J$ denote its set of at
most $r_q$ exceptional component indices; randomizing only these gives
$\sum_{j\in J}\lambda_j^2\leq r_q(K/n)^2$.  Taking the better construction
and paying $4\varepsilon$ for deletion proves
\eqref{eq:hyperfinite-convexification-l2}.

For the matching lower bound, the map above satisfies
$\|B(x)-B(y)\|_{\rm F}\geq2^{-1/2}\|x-y\|_2$.
If $N\geq2m$, take $x$ uniform on $2m$ type coordinates; its distance from
$\mathcal T_{N,m}$ is at least $1/\sqrt{2m}$.  If $N<2m$, take all
fractional parts equal to $s/N$ for $s=\lfloor N/2\rfloor$ as above; its
Euclidean distance from the grid is
$m^{-1}\sqrt{s(N-s)/N}=\Theta(\sqrt N/m)$.  Together with the upper bound,
these two cases prove
$\kappa_{q,2}(M_{2m})=\Theta(\min\{m^{-1/2},\sqrt N/m\})$.
\end{proof}

\subsubsection{Proof of Proposition~\ref{prop:projected-convexification}}
\label{app:proof-projected-convexification}

\begin{proof}
Let $H$, $C_1,\ldots,C_m$ and $\lambda_j=|V(C_j)|/n$ be as in the proof of
Theorem~\ref{thm:hyperfinite-convexification}.  Since $\Pi$ is linear,
applying it to \eqref{eq:component-minkowski-sum} gives
$\Pi\cS_q(H)=\sum_j\lambda_j\Pi\cS_q(C_j)$, and taking convex hulls (which
commute with linear images and with Minkowski sums) gives
$\operatorname{conv}\Pi\cS_q(H)=\sum_j\lambda_j\operatorname{conv}\Pi\cS_q(C_j)$.
The sets $\Pi\cS_q(C_j)$ lie in $\mathbb R^k$, so the weighted
Shapley--Folkman lemma of Appendix~\ref{app:deferred-task} applies in
dimension $k$ without any translation: every point of the hull equals
$\sum_j\lambda_ja_j$ with $a_j\in\Pi\cS_q(C_j)$ for all but at most $k$
indices $j$, the exceptional $a_j$ lying in
$\operatorname{conv}\Pi\cS_q(C_j)$.  Replacing each exceptional $a_j$ by
any point of $\Pi\cS_q(C_j)$ moves the sum by at most
$\lambda_j\operatorname{diam}_{\ell^1}\bigl(\operatorname{conv}\Pi\cS_q(C_j)\bigr)
 \leq\lambda_j\operatorname{diam}_{\ell^1}\bigl(\cS_q(C_j)\bigr)\leq\frac Kn\,2(1+D)$,
using that $\Pi$ is nonexpansive, that a convex hull has the diameter of
the set, and that $\mathsf Y_{D,q}$ has $\ell^1$-diameter at most
$2(1+D)$.  Hence
$d_{\rm H}(\Pi\cS_q(H),\operatorname{conv}\Pi\cS_q(H))\leq2(1+D)kK/n$.
The deletion step is as before: the same assignment on $G$ and on $H$
gives $d_{\rm H}(\cS_q(G),\cS_q(H))\leq2\varepsilon$, nonexpansiveness of
$\Pi$ gives the same bound for the projected sets, and taking hulls is
nonexpansive in Hausdorff distance; the triangle inequality proves
\eqref{eq:projected-convexification}.  The Lipschitz consequence is the
fact that minima (or maxima) of an $L$-Lipschitz function over two compact
sets differ by at most $L$ times their Hausdorff distance.

For sharpness, use the matching type map
\eqref{eq:matching-type-map}.  Select $k$ unordered color-pair types and one
additional type.  Give each selected type mass $1/(2m)$ and put the
remaining mass on the additional type.  This is possible when
$k\leq\min\{N_q-1,2m\}$.  Project onto the corresponding upper-triangular
$W$-coordinates.  An off-diagonal coordinate is a multiple of $1/(2m)$
for every realized matching quotient but equals $1/(4m)$ at the target;
a diagonal coordinate has the still larger distance $1/(2m)$ from its
$1/m$ grid.  Summing the $k$ coordinate distances proves
\eqref{eq:projected-convexification-lower}.
\end{proof}

\subsubsection{Proof of Proposition~\ref{prop:expander-separation}}
\label{app:proof-expander-separation}

\begin{proof}
The disjoint-union identity gives
$\cS_q(H_n)=\tfrac12\cS_q(G_n)+\tfrac12\cS_q(G_n)$, whose convex hull equals
$\operatorname{conv}\cS_q(G_n)$.  Coloring the two copies constantly with
different states gives $\Phi(H_n)=0$.  For a binary assignment on $G_n$, put
$x=p_1$ and $y=\min\{x,1-x\}$.  Its cut has size at least
$hy|V(G_n)|$, hence
\[
 |p_1-\tfrac12|+W_{12}
 \ge \tfrac12-y+hy\ge\min\{\tfrac12,\tfrac h2\}.
\]
\end{proof}

\subsubsection{Proof of Corollary~\ref{cor:necessity}}
\label{app:proof-necessity}

\begin{proof}
Let $(G_i)$ be $D$-regular graphs with $|V(G_i)|\to\infty$ and edge expansion
at least $h>0$; by Theorem~\ref{thm:compactness} pass to a
partition-convergent subsequence, still written $(G_i)$.  Alternate
$H_{2i-1}:=G_i$ and $H_{2i}:=G_i\sqcup G_i$.  By
\eqref{eq:expander-convex-equality},
$\operatorname{conv}\cS_q(H_{2i})=\operatorname{conv}\cS_q(G_i)$, so by
Proposition~\ref{prop:convex-sufficiency} every scalar energy converges along
the full sequence.  But the balanced-cut query
\eqref{eq:balanced-cut-query} is $1$-Lipschitz, hence Hausdorff-continuous,
and $\Phi(H_{2i})=0$ while
$\Phi(H_{2i-1})\geq\min\{1/2,h/2\}$, so $\cS_2(H_i)$ does not converge.
\end{proof}

\subsubsection{Proof of Proposition~\ref{prop:lagrangian-bridge-formal}}
\label{app:proof-lagrangian-bridge}

\begin{proof}
Weak duality: for $\|u\|_\infty\leq L$,
$\langle u,p-a\rangle\leq L\|p-a\|_1$ pointwise, so
$\mathsf g_{B,a}(u)\leq\mathsf m_{B,L,a}(G)$, proving the left inequality.

For the right inequality, write
$F(p,W):=\tfrac12\langle B,W\rangle+L\|p-a\|_1$ and
$\varphi\bigl((p,W),u\bigr):=\tfrac12\langle B,W\rangle+\langle u,p-a\rangle$
on $\operatorname{conv}\cS_q(G)\times\{u:\|u\|_\infty\leq L\}$.  Both
factors are compact and convex, and $\varphi$ is affine in each argument, so
the minimax theorem \citep{sion1958minimax} gives
\[
 \min_{z\in\operatorname{conv}\cS_q(G)}F(z)
 =\min_{z}\max_{u}\varphi(z,u)
 =\max_{u}\min_{z}\varphi(z,u)
 =\max_{\|u\|_\infty\leq L}\mathsf g_{B,a}(u),
\]
where the first equality uses
$L\|p-a\|_1=\max_{\|u\|_\infty\leq L}\langle u,p-a\rangle$ and the last uses
that an affine functional has the same minimum on $\cS_q(G)$ and its convex
hull.  Finally, $F$ is Lipschitz with constant
$\max\{L,\tfrac12\|B\|_{\max}\}$ for the metric $d_q$, because
$|\tfrac12\langle B,W\rangle-\tfrac12\langle B,W'\rangle|
 \leq\tfrac12\|B\|_{\max}\|W-W'\|_1$ and the penalty term is
$L$-Lipschitz in $\|p-p'\|_1$.  Every minimizer of $F$ over
$\operatorname{conv}\cS_q(G)$ lies within $\kappa_q(G)$ of $\cS_q(G)$, so
$\mathsf m_{B,L,a}(G)\leq\min_{\operatorname{conv}\cS_q(G)}F
 +\max\{L,\tfrac12\|B\|_{\max}\}\,\kappa_q(G)$.
\end{proof}

\subsection{Proofs for Section~\ref{sec:examples}}

The section defers the independence-ratio identity and the compact-spin
transfer proposition.
\subsubsection{Proof of the independence-ratio identity of Section~\ref{sec:examples}}

\begin{proof}
An assignment determines the selected set $S=\{v:c(v)=1\}$.  Its objective is
$(|S|-|E(G[S])|)/n$.  Choose one endpoint of each edge in the induced graph
$G[S]$, and let $C$ be the union of the chosen vertices.  The set $C$ covers
every edge of $G[S]$ and satisfies $|C|\leq|E(G[S])|$, so $S\setminus C$ is
independent.  Hence $|S|-|E(G[S])|\leq\alpha(G)$.  A maximum independent set
attains equality.
\end{proof}

\subsubsection{Proof of Proposition~\ref{prop:compact-spin}}
\label{app:proof-compact-spin}

\begin{proof}
Let $E(G):=\mathsf E^-_{K,h}(G)$.  Choose a finite $\delta$-net
$Z_\delta=\{z_1,\ldots,z_{q_\delta}\}\subseteq K$ and restrict all spins to
this net.  The resulting finite-state energy is
\begin{equation}
 E_\delta(G)=
 \min_{(p,W)\in\cS_{q_\delta}(G)}
 \frac12\sum_{i,j}h(z_i,z_j)W_{ij},
 \label{eq:finite-net-energy}
\end{equation}
so it extends continuously by
Theorem~\ref{thm:task-completeness}.

Let $\omega_h(\delta)$ be the largest change in $h$ when both arguments move
by at most $\delta$.  Restriction can only raise a minimum, while rounding a
minimizing $K$-assignment changes each edge term by at most
$\omega_h(\delta)$.  Because $|E(G)|/n\leq D/2$,
\begin{equation}
 0\leq E_\delta(G)-E(G)
 \leq\frac D2\,\omega_h(\delta).
 \label{eq:finite-net-uniform-error}
\end{equation}
It follows that, for two nets,
\begin{equation}
 |E_\delta(G)-E_\eta(G)|
 \leq\frac D2\bigl(\omega_h(\delta)+\omega_h(\eta)\bigr).
 \label{eq:finite-net-cauchy}
\end{equation}
Both finite-net energies are continuous on the limit space, and finite graph
states are dense, so the same inequality holds on the closure.  The family
$(E_\delta)$ is therefore uniformly Cauchy.  Its continuous limit is
independent of the nets and agrees with $E$ on every finite graph by
\eqref{eq:finite-net-uniform-error}.  Reversing the restriction inequality
gives the maximum case.
\end{proof}

\subsection{Proofs for Section~\ref{sec:size-transfer}}

The section defers the size-transfer theorem and its corollaries, the
encoder constructions with their treewidth lemma, the raw-graph RNI upper
and lower bounds, and the generalization theorem; the compact-spin transfer
corollary is stated as well as proved here.
\subsubsection{Proofs for task-value size transfer}
\label{app:proof-task-value-transfer}

\begin{proof}[Proof of Theorem~\ref{thm:task-value-size-transfer}]
By definition,
\begin{equation}
 \eta_{q,s}(G,G')=
 d_{{\rm H},s}\bigl(\operatorname{conv}\cS_q(G),
                      \operatorname{conv}\cS_q(G')\bigr).
 \label{eq:probe-convex-distance}
\end{equation}
For compact convex $C,C'$ in a normed space, support-function duality gives
$d_{\rm H}(C,C')=\sup_{\|\theta\|\leq1}|h_C(\theta)-h_{C'}(\theta)|$ over the
dual unit ball, so $\eta_{q,s}$ is exactly the discrepancy over dual-unit
scalar energies.  For $s=1$, our sets live in the subspace
$\{(p,W):W=W^\top\}$ with the $\ell^1$ norm
$\|p\|_1+\|W\|_1$, whose dual norm is
$\max\{\|a\|_\infty,\|B\|_{\max}\}$ over pairs $\theta=(a,B)$; restricting
$B$ to be symmetric loses no generality on this subspace, because for
symmetric $\Delta W$ one has
$\langle B,\Delta W\rangle=\langle\tfrac12(B+B^\top),\Delta W\rangle$ and
symmetrization does not increase $\|B\|_{\max}$.  The probe
$\ell_\theta$ in \eqref{eq:probe-energy} carries the factor $\tfrac12$ on
$B$, which is why $\|\theta\|_*$ uses $\tfrac12\|B\|_{\max}$: the map
$(a,B)\mapsto(a,\tfrac12B)$ matches $\ell_\theta$ with the dual pairing.
Replacing a coefficient by its negative converts minima to support
functions and identifies $\eta_{q,1}$ with $\eta_q$ in
\eqref{eq:scalar-probe-distance}.  Convexification is
nonexpansive, which proves the first inequality in
\eqref{eq:probe-quotient-sandwich}; the triangle inequality through the two
convex hulls proves the second.  Minima of an $L$-Lipschitz function over two
compact sets differ by at most $L$ times their Hausdorff distance, proving
\eqref{eq:lipschitz-task-transfer}.  Substituting
Theorem~\ref{thm:hyperfinite-convexification} gives
\eqref{eq:decomposable-size-transfer}, and substituting
\eqref{eq:hyperfinite-convexification-l2} gives the stated Euclidean
specialization.  If $F$ is affine, its minimum or maximum over a set
equals that over its convex hull, so the two outer $\kappa$ terms are
unnecessary.

For the finite-probe claim, $\|p\|_1+\|W\|_1\leq1+D$ on
$\mathsf Y_{D,q}$.  Approximating an arbitrary dual unit vector by
$\mathcal N_\rho$ changes each of the two probe values by at most
$(1+D)\rho$, which proves \eqref{eq:finite-probe-transfer}.
\end{proof}

\begin{proof}[Proof of Corollary~\ref{cor:opening-task-transfer}]
Each task is a single probe energy with an explicit dual norm:
$e_q=\gamma_\theta$ for $\theta=(0,I_q)$, since
$\ell_\theta=\tfrac12\sum_iW_{ii}=u_q$, with $\|\theta\|_*=\tfrac12$;
$\operatorname{defect}_J=\gamma_\theta$ for $\theta=(0,B)$ with
$B_{ij}=\mathbf 1[ij\notin E(J)]$, again $\|\theta\|_*=\tfrac12$;
$\operatorname{MaxCut}/n=-\gamma_\theta$ for $\theta=(0,B)$ with
$B_{12}=B_{21}=-1$ and zero diagonal, $\|\theta\|_*=\tfrac12$; and
$\alpha/n=-\gamma_\theta$ for $\theta=(a,B)$ with $a=(-1,0)$,
$B=\operatorname{diag}(1,0)$, $\|\theta\|_*=1$.  By homogeneity of
$\eta_q$ in \eqref{eq:scalar-probe-distance},
$|\gamma_\theta(G)-\gamma_\theta(G')|\leq\|\theta\|_*\,\eta_q(G,G')$,
which gives \eqref{eq:color-defect-transfer}--\eqref{eq:independence-transfer};
$\eta_q\leq\delta_q$ is the first inequality of
\eqref{eq:probe-quotient-sandwich}.  For the last claim, $e_q(G)=0$ and
$e_q(G')\leq\tfrac12\eta_q(G,G')$ means some assignment of $G'$ leaves at
most $\tfrac12\eta_q(G,G')|V(G')|$ monochromatic edges; deleting them
makes $G'$ $q$-colorable.
\end{proof}

\begin{proof}[Proof of Corollary~\ref{cor:finite-bank-size-generalization}]
The finite-graph representations form a totally bounded subset of the compact
finite product of hyperspaces, so they admit a finite $r$-net whose centers
are finite graph representations.  Their orders have a finite maximum
$n_{\max}(D,\mathbf q,r)$.  Choose $B$ within $r$ of a test graph $G$.
The target vector moves by at most $L_*r$, while
$d_R(\widehat R(G),\widehat R(B))\leq r+2\zeta$.  Inserting the prediction and
target at $B$ between those at $G$ gives
\eqref{eq:finite-bank-generalization-bound}.

For any $\xi>0$, choose a finite $\xi$-net $Z_\xi\subset K$, put
$m:=|Z_\xi|$, and let $\omega_h(\xi)$ be the supremum change in $h$ when
each argument moves by at most $\xi$.  For a compact-spin target, append the
finite-net coordinate $\cS_m$ to the representation.  The two rounding errors and
\eqref{eq:compact-spin-size-transfer} add $D\omega_h(\xi)$ to the corresponding
target-transfer term; for relaxed $d$-dimensional edge orthogonality this is
at most $4D\xi$.
\end{proof}

\subsubsection{Compact-spin transfer}
\label{app:proof-compact-spin-transfer}

\begin{corollary}[Compact-spin size transfer]
\label{cor:compact-spin-transfer}
For the compact-spin energy in Proposition~\ref{prop:compact-spin}, let
$Z_\xi\subset K$ be a finite $\xi$-net of size $m$, and let
$\omega_h(\xi)$ be the supremum change in $h$ when each argument moves by at
most $\xi$.  Then
\begin{equation}
 |\mathsf E^-_{K,h}(G)-\mathsf E^-_{K,h}(G')|
 \leq D\omega_h(\xi)+\frac12\|h\|_\infty\delta_m(G,G').
 \label{eq:compact-spin-size-transfer}
\end{equation}
For $K=S^{d-1}$ and $h(x,y)=\langle x,y\rangle^2$, one may take
$m\leq(1+2/\xi)^d$ and $\omega_h(\xi)\leq4\xi$, giving
$|\varepsilon_d^\perp(G)-\varepsilon_d^\perp(G')|
 \leq4D\xi+\tfrac12\delta_m(G,G')$.
\end{corollary}

The proof combines \eqref{eq:finite-net-uniform-error} with
\eqref{eq:lipschitz-task-transfer} applied to the finite-net energy
\eqref{eq:finite-net-energy}, whose integrand is
$\tfrac12\|h\|_\infty$-Lipschitz in $W$.

\subsubsection{Bounded treewidth: hyperfiniteness and exact quotient computation}

\begin{lemma}[Explicit hyperfiniteness of bounded treewidth]
\label{lem:treewidth-hyperfinite}
Let $G$ be $D$-bounded with treewidth at most $t$ and $n$ vertices.  For
every $K\geq1$, one can delete at most $2(t+1)Dn/K$ edges so that every
remaining component has at most $K$ vertices.
\end{lemma}

\begin{proof}
A graph of treewidth at most $t$ on $s$ vertices has a bag whose removal
leaves components of at most $s/2$ vertices each (a standard centroid-bag
argument on a tree decomposition).  Deleting the at most $(t+1)D$ edges
incident to that bag disconnects the bag vertices and leaves all parts of
order at most $s/2$.  Recurse on every part of order exceeding $K$.  Let
$T(s)$ be the maximum number of edges deleted for a piece of order $s$; we
claim $T(s)\leq(t+1)D(2s/K-1)$ for $s>K$, which gives the lemma.  Induct on
$s$: one split costs $(t+1)D$ and yields parts $s_1,s_2,\ldots$ with
$\sum_is_i\leq s$ and $s_i\leq s/2$.  Parts of order at most $K$ cost
nothing.  If no part exceeds $K$ the claim is immediate from $s>K$.  If
exactly one part exceeds $K$, then $s_1\leq s/2\leq s-K/2$, and
$(t+1)D\bigl(1+2s_1/K-1\bigr)\leq(t+1)D(2s/K-1)$.  If $\ell\geq2$ parts
exceed $K$, then
$(t+1)D\bigl(1+\sum_i(2s_i/K-1)\bigr)
 \leq(t+1)D(2s/K-1)$ using $\ell\geq2$.
\end{proof}

\begin{proof}[Proof of Proposition~\ref{prop:treewidth-encoder}]
Convert the given decomposition to a nice tree decomposition with introduce,
forget, join, and introduce-edge nodes, of the same width and $O((t+1)n)$ nodes;
each edge is processed at its unique introduce-edge node.
Scale coordinates by $n$: an assignment $c$ induces the integer vector
$\nu(c)=(np,nW)\in\mathbb Z^q\times\mathbb Z^{q\times q}$ with
$\sum_inp_i=n$ and $\sum_{ij}nW_{ij}=2|E(G)|\leq Dn$, so the number of
attainable vectors is at most $(n+1)^{q-1}(Dn+1)^{q(q+1)/2}=n^{O(q^2)}$.

For each node $x$ with bag $B_x$, and each coloring
$\sigma:B_x\to[q]$, let $\mathrm{DP}_x(\sigma)$ be the set of partial vectors
$\nu$ realizable by colorings of the vertices introduced strictly below $x$
(together with $\sigma$), where a vertex contributes its $p$-count when it is
forgotten and an introduce-edge node contributes its edge's two oriented
$W$-counts.  Leaves start from the empty vector.  An introduce-vertex node
extends $\sigma$ in $q$ ways; an introduce-edge node adds the corresponding
counts, which are determined by $\sigma$.  A forget node projects $\sigma$
and adds one $p$-count.  A join node combines the two children sets of the same $\sigma$ by vector
sums, a Minkowski sum of two subsets of the $n^{O(q^2)}$-point grid.  Each
node therefore costs at most $q^{t+1}\cdot n^{O(q^2)}$ set operations, and
the root returns exactly $\{\nu(c):c\}$, hence $\cS_q(G)$ after dividing by
$n$.  Claims (i) and (ii) follow because minima of
$\ell_\theta$ over the listed set are exact.

For (iii): apply Lemma~\ref{lem:treewidth-hyperfinite} with
$K:=\bigl\lceil2\sqrt{(t+1)Dn/((1+D)r_q)}\bigr\rceil$ and
Theorem~\ref{thm:hyperfinite-convexification} with
$\varepsilon=2(t+1)D/K$, giving
\[
 \kappa_q(G)\leq\frac{8(t+1)D}{K}+\frac{2(1+D)r_qK}{n}
 \leq8\sqrt{(1+D)r_q(t+1)D}\;n^{-1/2}+\frac{2(1+D)r_q}{n},
\]
where the last term accounts for the ceiling.  Since
$2(1+D)r_q/n\leq2(1+D)r_q\,n^{-1/2}$, both $\kappa_q(G)$ and
$\kappa_q(G')$ are at most
$\bigl(8\sqrt{(1+D)r_q(t+1)D}+2(1+D)r_q\bigr)$ times $n^{-1/2}$ and
$n'^{-1/2}$ respectively, and
\eqref{eq:effective-certificate} follows from
\eqref{eq:probe-quotient-sandwich} and \eqref{eq:finite-probe-transfer}.
\end{proof}
\subsubsection{Proof of Proposition~\ref{prop:explicit-cover}}
\label{app:proof-explicit-cover}

\begin{proof}
Write $R_0:=1+D$; every $z\in\mathsf Y_{D,q}$ has $\|z\|_1\leq R_0$.  Two
facts are used repeatedly.  (a)~If $A_1,\ldots,A_N\subset\mathsf Y_{D,q}$
and $\mu,\mu'$ are probability vectors, then
$d_{\rm H}\bigl(\sum_\tau\mu_\tau A_\tau,\sum_\tau\mu'_\tau A_\tau\bigr)
 \leq R_0\,\|\mu-\mu'\|_1$: map $\sum_\tau\mu_\tau a_\tau$ to
$\sum_\tau\mu'_\tau a_\tau$ with the same $a_\tau$, and symmetrically.
(b)~If $H$ is a disjoint union of connected graphs of order at most $K$,
with $\mu_\tau$ the fraction of vertices lying in components of type
$\tau\in\mathcal T_K$, then
$\operatorname{conv}\cS_q(H)=\sum_\tau\mu_\tau\operatorname{conv}\cS_q(\tau)$
(by \eqref{eq:component-minkowski-sum}, grouping isomorphic components and
using $A+A=2A$ for convex $A$), and
$d_{\rm H}(\cS_q(H),\operatorname{conv}\cS_q(H))\leq2R_0r_qK/|V(H)|$ by
\eqref{eq:component-convexification}.

Let $G\in\mathcal C$ have $n$ vertices, and let $H$ be the graph after the
$(\varepsilon,K)$-decomposition, so $d_{\rm H}(\cS_q(G),\cS_q(H))\leq2\varepsilon=r/4$.
If $n\leq n_{\max}$, take $B:=H\in\mathcal B_r$ and we are done.  Otherwise
let $\mu$ be the type proportions of $H$, and define
$m'_\tau:=\lceil\mu_\tau n_0/|\tau|\rceil$ and
$B:=\bigsqcup_\tau m'_\tau\,\tau$, of order
$n_B:=\sum_\tau m'_\tau|\tau|\in[n_0,\,n_0+N_KK]$, so $B\in\mathcal B_r$.
Its type proportions $\mu'_\tau=m'_\tau|\tau|/n_B$ satisfy
$0\leq\mu'_\tau-\mu_\tau n_0/n_B\leq|\tau|/n_B\leq K/n_B$, whence
$\|\mu'-\mu n_0/n_B\|_1\leq N_KK/n_B$, while
$\|\mu n_0/n_B-\mu\|_1=1-n_0/n_B\leq N_KK/n_B$; thus
$\|\mu-\mu'\|_1\leq2N_KK/n_0$.  By (b) applied to $H$ and to $B$, by (a),
and by the triangle inequality,
\[
 \begin{aligned}
 d_{\rm H}(\cS_q(G),\cS_q(B))
 &\leq\frac r4+\frac{2R_0r_qK}{n}
       +\frac{2R_0r_qK}{n_B}+\frac{2R_0N_KK}{n_0}\\
 &\leq\frac r4+\frac{4R_0K(r_q+N_K)}{n_0}
 \leq\frac r4+\frac r2\leq r.
 \end{aligned}
\]
using $n>n_{\max}\geq n_0$, $n_B\geq n_0$, and the definition of $n_0$.
For several state spaces the same $B$ serves all $q_j$ at once because
$\mathcal T_K$ and $\mu$ do not depend on $q$; only $r_q$ does, and it is
monotone in $q$.  The bound $N_K\leq\sum_{k\leq K}2^{\binom k2}\leq K2^{K(K-1)/2}$
counts labeled graphs.
\end{proof}

\subsubsection{Proof of Lemma~\ref{lem:local-rni-compilation}}

\begin{proof}
The proof separates the finite local computation from its neural
implementation.  Absorb the fixed simulation and flooding overhead into a
constant-factor enlargement of $Q$; we keep the same symbol, so the resulting
depth remains $O(Q)$.  Every transcript of $\mathcal A$ visits at most $Q+1$
vertices in the radius-$Q$ ball of its root.  For each of the two rank
coordinates, divide $[0,1]$ into $M$ ordered bins and replace the hard bin
indicators by continuous piecewise-linear soft indicators.  They are exactly
one-hot outside narrow transition strips around the $M-1$ thresholds and are
implemented by a fixed ReLU multilayer perceptron.  Since a radius-$Q$ ball has at most
$b_D(Q):=1+D\sum_{j=0}^{Q-1}(D-1)^j$ vertices, a union bound makes the
probability of a bin collision in either coordinate at most $b_D(Q)^2/M$.
Choose $M$ so that this is at most $\zeta\xi/(16B)$, then choose the strips
with sufficiently small total measure that the probability some mark in the
ball meets a strip is at most the same quantity.  The expected fraction of
bad roots is therefore at most $\zeta\xi/(8B)$, and Markov's inequality makes
that fraction at most $\xi/(4B)$ with probability at least $1-\zeta/2$.

Outside those roots, the soft indicators are one-hot and the two quantized
ranks preserve every comparison and otherwise arbitrary enumeration order used by
$\mathcal A$.  There are only finitely many degree-$D$, depth-$Q$
marked transcripts.  A sum-aggregation MPNN can encode their finite
multisets injectively with one-hot states, flood the transcript to the root
in $O(Q)$ rounds, and implement its finite output table by a multilayer
perceptron; equivalently, this is the bounded-order RNI construction of
\citet{abboud2021surprising,sato2021random} applied only to the queried
transcript.  Approximate the finite table to rootwise error at most
$\xi/2$ and clip the output to $[-B,B]^P$.  Good roots then contribute at
most $\xi/2$ to the vertex mean and bad roots at most
$2B\cdot\xi/(4B)=\xi/2$.  Adjusting the two probability allocations proves
\eqref{eq:local-rni-compilation}.  All alphabets, widths, and weights depend
on $(D,Q,P,B,\xi,\zeta)$ but not on $n$.
\end{proof}

\subsubsection{Proof of Theorem~\ref{thm:raw-rni-encoder}}
\label{app:proof-raw-rni-encoder}

\begin{proof}
We first invoke the bounded-treewidth partition oracle, compile its rooted
queries with Lemma~\ref{lem:local-rni-compilation}, and then amplify the
constant success probability by independent median channels.

The partition oracle of \citet[Theorem~1]{edelman2011partitioning} uses iid
priority ranks for its random seed.  Give each vertex a second independent
iid rank and use it to order every otherwise arbitrary vertex enumeration,
including loops over discovered vertex sets.  Conditional on these auxiliary
ranks the enumeration order is query-independent, as required
in that theorem, while the cut-probability proof over the priority ranks is
unchanged.  The oracle gives query access to a consistent partition
whose parts have order at most $K_\alpha$, cuts at most $\alpha n$ edges
with probability at least $9/10$, and uses at most $Q_\alpha$ adjacency
queries per root, with the bounds in
\eqref{eq:partition-oracle-complexity}.  Another $O(DK_\alpha)$ queries
recover all edges induced by the returned part; this is absorbed into the
stated depth.

For a root $v$, let $C(v)$ be its oracle part and let the local algorithm
return the vector $(\gamma_\theta(C(v)))_{\theta\in\Theta}$.  It is bounded
by $1+D$, comparison based, local, and equivariant.  Invoke Lemma~
\ref{lem:local-rni-compilation} with $\zeta=1/10$ to obtain one RNI-MPNN
channel with vertex-mean error at most $\xi$ with probability at least
$9/10$.  If $H$ is the graph after the oracle cut,
affinity of each probe and the disjoint-union identity give
\[
 \frac1n\sum_v\gamma_\theta(C(v))
 =\sum_{C\in H}\frac{|C|}{n}\gamma_\theta(C)
 =\gamma_\theta(H).
\]
The same assignment before and after deletion shows
$d_{\rm H}(\cS_q(G),\cS_q(H))\leq2\alpha$, so probe duality yields
$|\gamma_\theta(G)-\gamma_\theta(H)|\leq2\alpha$ simultaneously for the
finite bank.

By a union bound with the oracle's $9/10$ event, one channel therefore
satisfies the target inequality with probability at least $4/5$.  Run
$T=O(\log(1/\delta))$ independent channels and take the coordinatewise
median.  A Chernoff bound makes a strict majority of channels successful
with probability at least $1-\delta$, proving
\eqref{eq:raw-rni-probe-error}.  Parallel channels change width, not depth,
and no parameter depends on graph order.
\end{proof}

\subsubsection{Proof of Theorem~\ref{thm:rni-depth-lower}}
\label{app:proof-rni-depth-lower}

\begin{proof}
We separate a high-girth graph from its bipartite double cover by coloring
defect, identify their depth-$t$ local laws below the girth scale, and use
replication to concentrate their mean-pooled representations.

Take a nonbipartite $D$-regular Lubotzky--Phillips--Sarnak Ramanujan family
$X_K$ with girth $g_K\geq c\log K$ for some constant $c>0$ and
$\lambda_{\min}(X_K)\geq-2\sqrt{D-1}=-D+\delta_D$, where
$\delta_D:=D-2\sqrt{D-1}>0$ \citep{lubotzky1988ramanujan}.  Let $B_K$ be
the bipartite double cover of $X_K$ and define
$A_{K,L}:=L B_K$ and $H_{K,L}:=2L X_K$.  Both graphs have $2KL$ vertices
and component order at most $2K$.  Bipartiteness gives $e_2(A_{K,L})=0$.
For a binary coloring of $X_K$, encoded by $x\in\{-1,1\}^K$, the number of
monochromatic edges divided by $K$ is
\[
 \frac{D+x^\top A_{X_K}x/K}{4}\geq\frac{\delta_D}{4}=:c_0,
\]
which proves the other half of \eqref{eq:rni-depth-separation}.

Suppose $2t+1<g_K$.  Every radius-$t$ unmarked rooted neighborhood in both
graphs is the same $D$-regular tree.  With iid initialization, the final
node state of any $t$-layer local MPNN therefore has the same distribution
and mean $\mu$ on both graphs.  States at vertices more than $2t$ apart
depend on disjoint initializations.  If $\bar h$ is the vertex mean and
the state lies in $[-M,M]^w$, the dependency-graph variance bound gives
\begin{equation}
 \mathbb E\|\bar h-\mu\|_2^2
 \leq\frac{4M^2w\,b_D(2t)}{2KL}.
 \label{eq:rni-mean-concentration}
\end{equation}
For a $\Lambda$-Lipschitz readout, the expected difference between its
outputs on $A_{K,L}$ and $H_{K,L}$ is consequently at most
$4\Lambda M\sqrt{w b_D(2t)/(2KL)}$.  If both expected task errors are at
most $\eta$, the triangle inequality forces
\[
 c_0\leq2\eta+
 4\Lambda M\sqrt{\frac{w b_D(2t)}{2KL}}.
\]
For fixed finite $(w,M,\Lambda,t)$ and $\eta<c_0/2$, the rightmost term
vanishes as $L\to\infty$, a contradiction.  Hence a single network valid
for every repetition count must satisfy $2t+1\geq g_K$, and therefore
$t\geq(c/2)\log K-1/2$; after discarding finitely many $K$, take
$c_1=c/4$.  Global communication, deterministic positional
identifiers, sum pooling, or graph-size-dependent bounds invalidate the
locality or concentration step and are outside the theorem.
\end{proof}

\subsubsection{Proofs of Theorem~\ref{thm:generalization} and Propositions~\ref{prop:linear-generalization} and~\ref{prop:probe-generalization}}
\label{app:proof-generalization}

\begin{proof}
\emph{Covering the state space.}  If $\mathcal P$ is a $\rho$-net of the compact
metric space $Y$ and $A\subseteq Y$ is nonempty compact, then
$A_{\mathcal P}:=\{x\in\mathcal P:\operatorname{dist}(x,A)\leq\rho\}$ is nonempty and
$d_{\rm H}(A,A_{\mathcal P})\leq\rho$; hence the subsets of $\mathcal P$ form a $\rho$-net of
$\cK(Y)$ and $\mathsf N_{\cK(Y)}(\rho)\leq2^{|\mathcal P|}$.  Products with the max
metric multiply nets, giving
$\mathsf N_{\mathfrak X}(\rho)\leq\prod_j2^{\mathsf N_{Y_j}(\rho)}$, where
$Y_j:=\mathsf Y_{D,q_j}$, and the
volumetric bound for the $\ell^1$-ball of radius $1+D$ in dimension
$q_j+q_j^2$ gives $\mathsf N_{Y_j}(\rho)\leq(3(1+D)/\rho)^{q_j+q_j^2}$ for
$\rho\leq1+D$.

\emph{Uniform bound.}  Fix a $\rho$-net $P_{\mathfrak X}$ of
$\mathfrak X_s$ of size $\mathsf N_{\mathfrak X}(\rho)$ and round function
values to the grid $\{0,\Lambda\rho,2\Lambda\rho,\ldots\}\cap[0,B]$.  Two
$\Lambda$-Lipschitz functions sharing rounded values on
$P_{\mathfrak X}$ differ by at most $3\Lambda\rho$ in sup norm, so the class
admits a $3\Lambda\rho$-net of cardinality at most
$(1+B/(\Lambda\rho))^{\mathsf N_{\mathfrak X}(\rho)}$.  For each
representative, Hoeffding's inequality applies to the $[0,B]$-valued loss;
a union bound over representatives and the $2\times3\Lambda\rho$ transfer
between a function and its representative (once in the population term,
once in the empirical term) give \eqref{eq:generalization-bound} with the
stated logarithm since $1+u\leq3+u$.

\emph{Probe representation (Proposition~\ref{prop:probe-generalization}).}
Repeat the covering argument with the compact state space replaced by the
cube $[-(1{+}D),1{+}D]^P$ in $\|\cdot\|_\infty$: a grid of spacing $2\rho$
along each axis is a $\rho$-net with at most
$\lceil(1+D)/\rho\rceil^P\leq(2(1+D)/\rho)^P$ points for $\rho\leq1+D$,
and $\Lambda$-Lipschitz $[0,B]$-valued functions on the cube are covered
at scale $3\Lambda\rho$ by at most $(1+B/(\Lambda\rho))^{\mathsf N_P(\rho)}$
representatives exactly as before; Hoeffding, the union bound and the
$2\times3\Lambda\rho$ transfer give \eqref{eq:probe-generalization}.

\emph{Linear readouts (Proposition~\ref{prop:linear-generalization}).}  The loss class
$\{G\mapsto|f(G)-y(G)|:f\in\mathcal F_{\rm lin}\}$ takes values in
$[0,B_*]$ with $B_*:=W_1(1{+}D)+2B$ and is a $1$-Lipschitz image of
$\mathcal F_{\rm lin}$.  Let $\widehat{\mathfrak R}_m$ denote empirical
Rademacher complexity.  After scaling by $B_*$, apply the one-sided empirical
Rademacher bound \citep[Theorem~3.3]{mohri2018foundations} to the loss class
and to the class obtained by replacing each loss $h$ with $B_*-h$, each with
failure probability $\delta/2$.  Their empirical complexities agree by sign
symmetry, and Talagrand contraction bounds either by that of
$\mathcal F_{\rm lin}$.  With
probability $1-\delta$ the uniform gap is at most
$2\widehat{\mathfrak R}_m(\mathcal F_{\rm lin})
+3B_*\sqrt{\ln(4/\delta)/(2m)}$.  Finally,
$\widehat{\mathfrak R}_m(\mathcal F_{\rm lin})
\leq W_1(1{+}D)\sqrt{2\ln(2P)/m}+B/\sqrt m$: the first term is Massart's
lemma for the $\ell^1$-ball acting on features bounded by $1+D$ in sup
norm, and the second is the Rademacher complexity of the constant class
$\{b:|b|\leq B\}$.
\end{proof}

\subsection{Proofs for Appendix~\ref{sec:boundary}}

The repair-and-exactness appendix defers the edit-stability proposition, the
approximate-chromatic-number proposition, and the exact-colorability
theorem.
\subsubsection{Proof of Proposition~\ref{prop:disjoint-union}}
\label{app:proof-disjoint-union}

\begin{proof}
An assignment on a disjoint union is exactly a pair of assignments on its
components.  Both $p$ and $W$ are their vertex-weighted averages, which proves
\eqref{eq:disjoint-union}.  The diameter of each quotient host is at most
$2(1+D)$.  With
$\lambda_n:=|V(G_n)|/(|V(G_n)|+|V(H_n)|)$, the $q$th Hausdorff distance
after adding $H_n$ is at most
$2(1+D)(1-\lambda_n)$.  Dominated convergence in the product metric proves
\eqref{eq:sublinear-vertex-edit}.

For graphs on the same vertex set, use the same assignment in both graphs.
The node weights agree, and every edge in the symmetric difference changes
the $\ell^1$ mass of the oriented edge matrix by $2/|V(G_n)|$.  Hence each
$q$th Hausdorff distance is at most
$2|E(G_n)\triangle E(G_n')|/|V(G_n)|$.
\end{proof}

\subsubsection{Proof of Proposition~\ref{prop:approximate-chromatic-number}}
\label{app:proof-approximate-chromatic-number}

\begin{proof}
The defects decrease with $q$, and
$\{x:\chi_{\rm app}(x)>q\}=\{x:e_q(x)>0\}$ is open because $e_q$ is
continuous.  This proves lower semicontinuity.  On a finite graph,
$e_q(G)=0$ exactly when $G$ is $q$-colorable.  The remaining statements follow
from continuity and nonnegativity: if $\chi(G)=r$, only the $r-1$ terms with
$q<r$ can contribute to \eqref{eq:continuous-chromatic-lower-bound}.
\end{proof}

\subsubsection{Proof of Theorem~\ref{thm:exact-boundary}}
\label{app:proof-exact-boundary}

\begin{proof}
Let $I_n$ be the edgeless graph on $n$ vertices.  Proposition
\ref{prop:disjoint-union} gives the same partition limit for $I_n$ and
$I_n\sqcup K_{q+1}$, although only the first graph is $q$-colorable.  Hence no
continuous extension of the exact-colorability indicator exists.

For the upper-bound claim, take any growing sequence $G_n\to x$.  The graphs
$G_n\sqcup K_{D+1}$ have the same limit and chromatic number $D+1$.
Continuity of $U$ and its finite-graph upper-bound property imply
$U(x)\geq D+1$.
\end{proof}

\end{document}